\documentclass[10pt, letter, onecolumn]{arxiv}

\usepackage{helvet}
\usepackage{kantlipsum, lipsum}
\usepackage{dm-colors}
\usepackage{amsmath}
\usepackage{pstricks, pst-node}
\usepackage{verbatim}
\usepackage{multirow}
\usepackage{array}
\usepackage{longtable}
\usepackage{pdflscape}
\usepackage{scalerel}
\usepackage{booktabs}
\usepackage{enumitem}
\usepackage{xspace}
\usepackage{bm}
\usepackage{bbm}
\usepackage{mathtools}
\usepackage{soul}
\usepackage{epsfig}
\usepackage{graphicx}
\usepackage{amssymb}
\usepackage{colortbl}
\usepackage{csquotes}
\usepackage{setspace}
\usepackage{bbding}
\usepackage{siunitx} % 用于对齐百分比
\usepackage{threeparttable}
\usepackage{tabularx,ragged2e}
\usepackage{placeins}
\usepackage{bbding}
\usepackage{subcaption}
\definecolor{darkpastelgreen}{rgb}{0.13, 0.55, 0.13}
\definecolor{darkpastelred}{rgb}{0.55, 0.13, 0.13}
\definecolor{mygray}{rgb}{1, 1, 1}
\usepackage{lmodern}
\usepackage[hang,flushmargin]{footmisc}

\usepackage{nameref}
\usepackage{varioref}
\usepackage{amssymb}% http://ctan.org/pkg/amssymb
\usepackage{pifont}% http://ctan.org/pkg/pifont
\usepackage{rotating}
\usepackage{graphicx}

\usepackage[pagebackref=false,breaklinks=false,%
            colorlinks=true,bookmarks=true,citecolor=ourdarkblue,%
            urlcolor=ourdarkblue,linkcolor=ourdarkblue]{hyperref}
\usepackage[noabbrev,capitalize]{cleveref}
\usepackage{etoc}
\usepackage{lineno}
\usepackage{tikz}

\usepackage{wasysym}
\usepackage{algorithm}
\usepackage{makecell} 
\usepackage{algpseudocode}
\usepackage{amsmath}
\usepackage{thmtools}
\usepackage{tcolorbox}
\usepackage{wasysym}
\declaretheoremstyle[
    spaceabove=6pt, spacebelow=6pt,
    headfont=\bfseries, headpunct={.}, headformat={\NAME\ \NUMBER},
    bodyfont=\normalfont,
    postheadspace=0.5em
]{promptstyle}

\tcolorboxenvironment{prompt}{
    colback=gray!10!,
    colframe=gray!75!,
    fonttitle=\bfseries,
    title=Prompt,
    boxrule=0.5pt,
    sharp corners
}

\graphicspath{{figures/}}
\definecolor{mygray}{rgb}{0.85, 0.85, 0.85}

\usepackage{listings}
\usepackage{xcolor}

\definecolor{codegreen}{rgb}{0,0.6,0}
\definecolor{codegray}{rgb}{0.5,0.5,0.5}
\definecolor{codepurple}{rgb}{0.58,0,0.82}
\definecolor{backcolour}{rgb}{0.95,0.95,0.92}
\definecolor{framecolor}{rgb}{0.8,0.8,0.8}

\lstdefinestyle{prettyjson}{
    backgroundcolor=\color{backcolour},   
    commentstyle=\color{codegreen},
    keywordstyle=\color{blue}\bfseries,
    numberstyle=\tiny\color{codegray},
    stringstyle=\color{codepurple},
    basicstyle=\ttfamily\small,
    breakatwhitespace=false,         
    breaklines=true,                 
    captionpos=b,                    
    keepspaces=true,                 
    numbers=left,                    
    numbersep=8pt,                  
    showspaces=false,                
    showstringspaces=false,
    showtabs=false,                  
    tabsize=2,
    frame=single,
    frameround=tttt,
    framerule=0.5pt,
    rulecolor=\color{framecolor},
    xleftmargin=15pt,
    xrightmargin=15pt,
    aboveskip=15pt,
    belowskip=15pt,
    columns=flexible,
    escapeinside={(*@}{@*)}
}

\title{\Large{Teaching AI Stepwise Diagnostic Reasoning with Report-Guided Chain-of-Thought Learning}}

\author[1,*]{Yihong Luo} %#1 $\ast$
\author[1,2,*]{Wenwu He} %#1 $\ast$
\author[3,4,$\dagger$]{Zhuo-Xu Cui}
\author[3,4,$\dagger$]{Dong Liang}

\affil[1]{\normalsize Fujian University of Technology, Fuzhou, China \authorcr \vspace{0.1cm}}

\affil[2]{\normalsize Fujian Provincial Key Laboratory of Big Data Mining and Applications, Fuzhou,
China \authorcr \vspace{0.1cm}
}

\affil[3]{\normalsize Shenzhen Institute of Advanced Technology Chinese Academy of Sciences, Shenzhen, China \authorcr \vspace{0.1cm}
}

\affil[4]{\normalsize Key Laboratory of Biomedical Imaging Science and System, Chinese Academy of
 Sciences, Shenzhen, China \authorcr \vspace{0.1cm}
}

\affil[$\ast$]{\normalsize Equal contributions\hspace{1cm}}
\affil[$\dag$]{\normalsize Corresponding author\authorcr Zhuo-Xu Cui: zx.cui@siat.ac.cn; Dong Liang: dong.liang@siat.ac.cn}

\begin{document}
\begin{abstract}

This study presents DiagCoT, a multi-stage framework that applies supervised fine-tuning to general-purpose vision–language models (VLMs) to emulate radiologists’ stepwise diagnostic reasoning using only free-text reports. DiagCoT combines contrastive image–report tuning for domain alignment, chain-of-thought supervision to capture inferential logic, and reinforcement tuning with clinical reward signals to enhance factual accuracy and fluency. On the MIMIC-CXR benchmark, DiagCoT improved zero-shot disease classification AUC from 0.52 to 0.76 (absolute gain of 0.24), pathology grounding mIoU from 0.08 to 0.31 (absolute gain of 0.23), and report generation BLEU from 0.11 to 0.33 (absolute gain of 0.22). It outperformed state-of-the-art models including LLaVA-Med and CXR-LLAVA on long-tailed diseases and external datasets. By converting unstructured clinical narratives into structured supervision, DiagCoT offers a scalable approach for developing interpretable and diagnostically competent AI systems for radiology.
\end{abstract}

\maketitle
% ------------ SECTIONS ---------------------
% ========== Edit your name here
\section{Introduction}

Vision–language models (VLMs)~\cite{liu2023visual,bai2023qwenvlversatilevisionlanguagemodel,liu2024deepseek,wang2024cogvlm} have recently transformed a wide range of general-domain multimodal tasks, including visual question answering, image captioning, and multimodal dialogue, driven by large-scale image–text pre-training and supervised fine-tuning. This success has stimulated growing interest in applying VLMs to medical imaging, particularly radiology, where large-scale paired image–report datasets offer rich aligned visual and textual information~\cite{monshi2020deep}. However, unlike natural images, clinical interpretation requires not only accurate pattern recognition but also hierarchical clinical reasoning, adherence to expert consensus, and precise use of specialized medical terminology aligned with diagnostic workflows.

Early automated radiology report generation methods leveraged convolutional encoder–recurrent decoder architectures enhanced by attention mechanisms to connect visual and textual modalities~\cite{2015Deep,2015Long,2015Show,2018Bottom}. While promising, these models often struggled to capture subtle pathological features and generated narratives lacking clinical coherence and depth. The adoption of Transformer-based architectures has since significantly advanced image encoding and language generation, resulting in more context-aware and coherent outputs. Yet, a substantial domain gap remains: VLMs pretrained on natural images and general text lack the anatomical priors, clinical reasoning frameworks, and terminological accuracy necessary for reliable diagnostic interpretation in real-world medical settings.

\begin{figure*}[t!]
\begin{center}
\includegraphics[width=\textwidth]{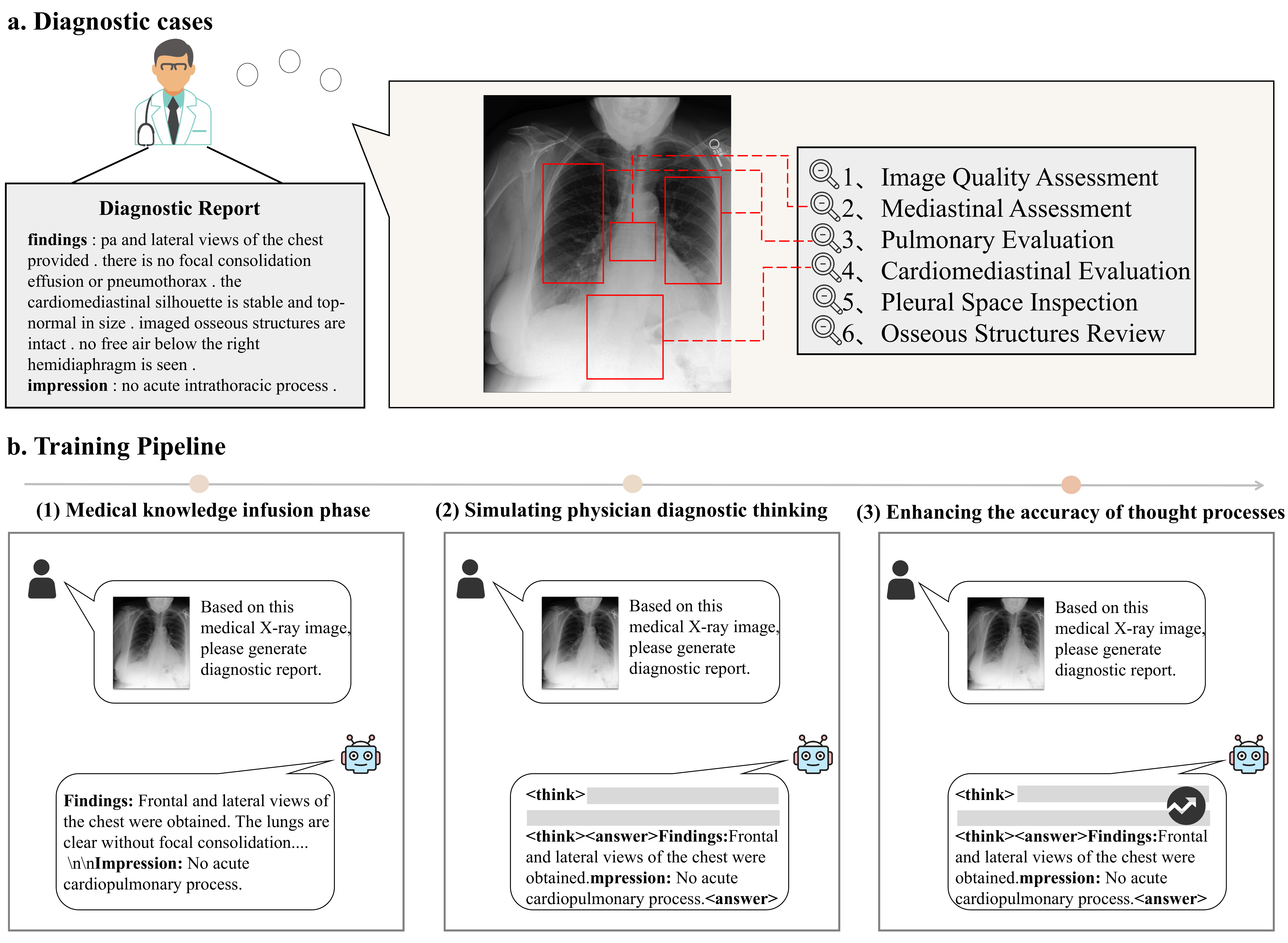}
\end{center}
\caption{\textbf{Overview of DiagCoT.} (a). It illustrates the process by which a radiologist evaluates medical images from six distinct perspectives to formulate a final diagnostic report. (b). It delineates the DiagCoT training pipeline, which is structured into three primary phases. The first phase, the Medical Knowledge Infusion Phase, aims to equip the model with the foundational ability to generate preliminary reports based on medical images. The second phase, the Simulating Physician Diagnostic Thinking Phase, is designed to train the model to emulate the diagnostic reasoning process of radiologists, thereby yielding accurate diagnostic reports. The third phase, the Enhancing the Accuracy of Thought Processes Phase, seeks to refine the reasoning processes acquired in the second phase by employing a reinforcement learning algorithm, ultimately enhancing the model's diagnostic capabilities.}
%(c). This figure presents the performance of DiagCoT on the MIMIC-CXR test dataset. The results demonstrate that DiagCoT achieves superior performance on the BLEU-\{1, 2, 3, 4\} metrics compared to other baseline models.}
   \label{fig: overview}
  % \vspace{-.15in}
\end{figure*}

A critical and underexplored challenge is the accurate diagnosis of rare thoracic diseases such as pneumomediastinum, pulmonary fibrosis, and lymphangitic carcinomatosis. These conditions are characterized by low prevalence and often present with subtle or atypical radiographic signs, which contribute to high rates of misdiagnosis or delayed detection. Since VLMs tend to learn predominantly from common disease patterns, their ability to generalize to these rare, long-tail categories is limited, particularly under conditions of sparse supervision. Overcoming this barrier requires methods that combine data efficiency with explicit clinical reasoning to interpret ambiguous or low-frequency presentations.

To address these limitations, Referring to Figure~\ref{fig: overview}, we propose DiagCoT, a multi-stage fine-tuning framework that endows VLMs with diagnostic reasoning abilities through structured learning from paired chest X-rays and expert-authored radiology reports. DiagCoT sequentially implements: (i)  Medical knowledge infusion phase to establish precise spatial and semantic grounding between visual features and textual descriptions; (ii) Simulating physician diagnostic thinking using chain-of-thought (CoT)~\cite{wei2022chain} tuning to embed intermediate reasoning steps reflecting clinical decision-making processes; and (iii)  Enhancing the accuracy of thought processes using reinforcement optimization to refine factual accuracy and linguistic fluency by rewarding semantically and clinically valid outputs.

Unlike prior approaches treating reports as mere outputs, DiagCoT leverages them as rich instructive signals to internalize medical knowledge and reasoning pathways that transfer effectively across multiple downstream tasks. This design not only enhances performance in disease classification, pathology grounding, and report generation but also notably improves generalization to rare and underrepresented disease categories. Additionally, we investigate augmenting rare disease representation through text-guided generative models that synthesize realistic chest X-ray images, further boosting model robustness in low-data regimes.

By integrating domain-specific knowledge acquisition with explicit reasoning mechanisms, DiagCoT offers a foundational paradigm for developing clinically reliable, interpretable, and task-adaptive multimodal AI models. This framework has strong potential to accelerate AI deployment in routine medical imaging workflows—especially for conditions where diagnostic uncertainty is greatest and expert resources are limited—ultimately enhancing diagnostic accuracy and improving patient outcomes.

\section{Results}
\subsection{Implementation}
\textbf{Datasets.} For X-Ray Report Generation Within-dataset Evaluation, Our method primarily uses the MIMIC-CXR~\cite{johnson2019mimic} dataset, containing over 220,000 chest X-ray images from more than 65,000 patients. Training data is approximately 360,000 entries (frontal/lateral views share one reference report). Due to missing Findings or Impression sections in some reports, data preprocessing is needed, leaving about 220,000 training samples and 2,580 test samples. For X-Ray Report Generation Cross-dataset Evaluation, we use IU-Xray~\cite{demner2015preparing} dataset, which consists of 3955 reports and 7,470 chest X-ray images, a random selection of 1,180 images was drawn from the dataset for testing purposes.

\textbf{Metrics.} The following metrics assess predictive reports: BLEU\{1-4\}~\cite{papineni2002bleu}, ROUGE-L~\cite{lin2004rouge}, Meteor~\cite{denkowski2011meteor}, CIDEr~\cite{vedantam2015cider}.

\textbf{Our methods.} We detail models used:
Baseline model(Qwen2-VL-7B-Instruct~\cite{bai2023qwenvlversatilevisionlanguagemodel}), 
Teacher-VLM(Qwen2.5-VL-32B-Instruct~\cite{bai2023qwenvlversatilevisionlanguagemodel}), Optimizer-VLM(Qwen2.5-VL-72B-Instruct-AWQ~\cite{bai2023qwenvlversatilevisionlanguagemodel}), These Qwen2/2.5 series are efficient multimodal large models achieving vision-language alignment via hybrid encoding architecture and multi-stage cross-modal training, supporting long-context input.
During medical knowledge alignment, the baseline model is trained to gain basic report generation ability. In CoT-tuning stage, three models are used: the stage1-aligned model as $\mathcal{M}_\text{Stage1}$, Teacher-VLM as $\mathcal{M}_\text{med init}$, and Optimizer-VLM as $\mathcal{M}_\text{expert}$. For CoT dataset construction, LoRA fine-tunes the medical report-aligned baseline model, then collects high-quality CoT data. Generated data undergoes expert model evaluation to produce the final CoT dataset. For RFT-tuning stage, Based on the Stage2 model $\mathcal{M}_\text{Stage2}$, the GRPO algorithm is introduced to optimize the report generation task, aiming to significantly enhance the accuracy and reliability of the output reports. \\
\textbf{Baseline methods.} Our vision-language model is compared against leading medical report generation VLMs (e.g., LLaVA-Med~\cite{li2023llava}, CXR-LLAVA~\cite{lee2025cxr},ChestX-Reasoner~\cite{fan2025chestx}, CheXagent-2-3b~\cite{chen2024chexagent}, Deepseek-VL-7B-chat~\cite{lu2024deepseek}), the largest Qwen2.5-VL-72B-AWQ~\cite{bai2023qwenvlversatilevisionlanguagemodel} model, and our stage-aligned models. In addition, there are traditional medical report generation models, such as R2Gen~\cite{chen-emnlp-2020-r2gen}, R2GenCMN~\cite{chen-acl-2021-r2gencmn}, XproNet~\cite{wang2022cross}, and M2KT~\cite{yang2023radiology}.

\subsection{Main Results}
First, to validate the efficacy of DiagCoT for X-ray image diagnosis, we conducted comprehensive experiments encompassing X-Ray Report Generation (RRG), Classification, and Grounding tasks. Specifically for the RRG task, we implemented within-dataset evaluation and cross-dataset evaluation protocols to rigorously assess the model's generalization capabilities. Subsequently, both qualitative and quantitative results are presented to demonstrate DiagCoT's performance across report generation, classification, and grounding tasks.

\begin{figure*}[htbp]
\begin{center}
\includegraphics[width=\textwidth]{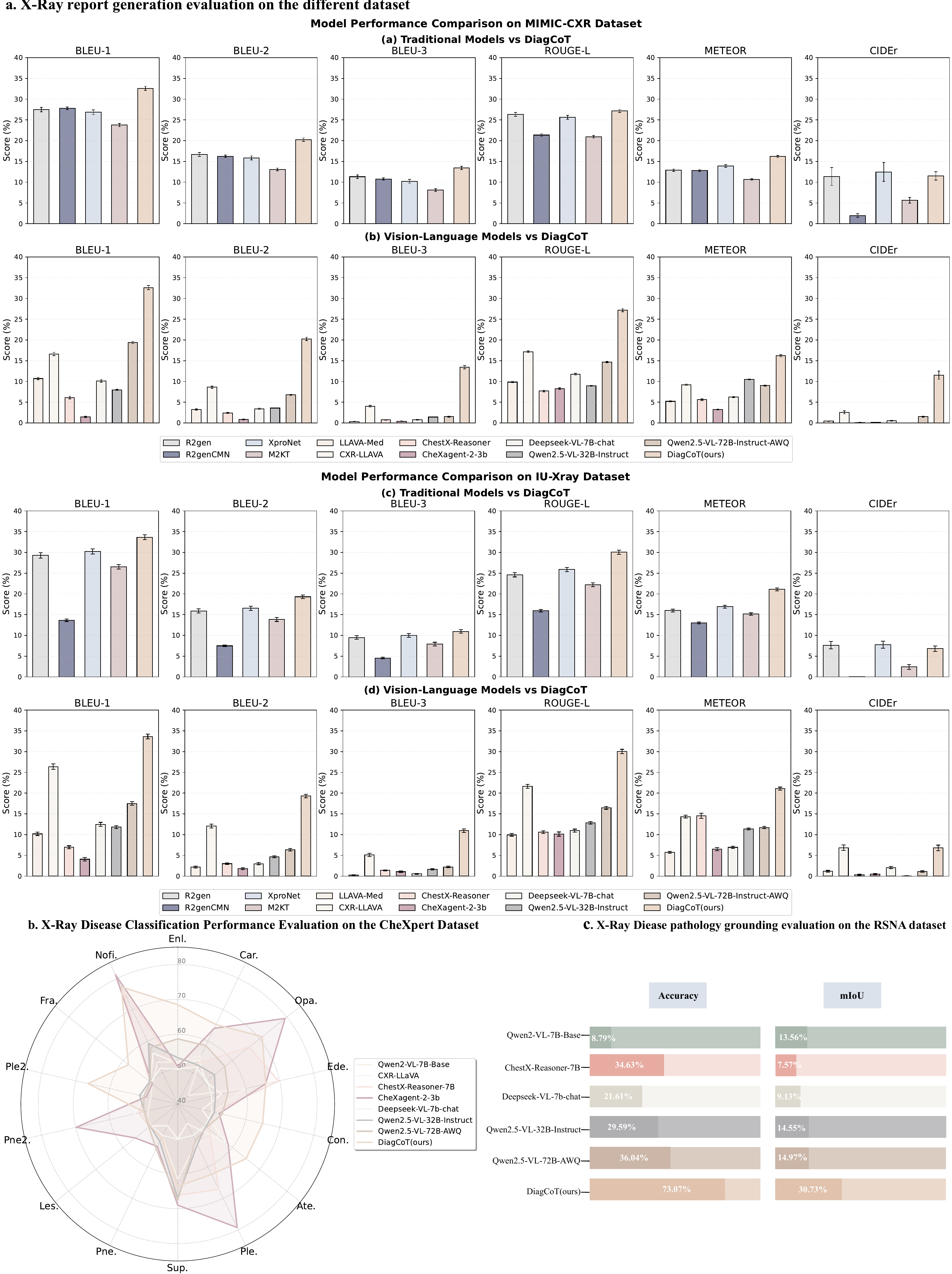}
\end{center}

\caption{\textbf{Results of DiagCoT compared with baseline models across different tasks.} a. X-Ray report generation evaluation on different datasets shows that DiagCoT (ours) outperforms both traditional models and vision-language models across metrics. b. Evaluation on the CheXpert dataset for X-Ray disease classification demonstrates the competitive performance of DiagCoT (ours) among compared models.
 c. X-Ray disease pathology grounding evaluation on the RSNA dataset. }
   \label{fig: Result_visual_all}
  % \vspace{-.15in}
\end{figure*}

%including BLEU\{1-3\}, ROUGE-L, METEOR, and CIDEr on both MIMIC-CXR and IU-Xray datasets,  
%, assessed via accuracy and mIoU, demonstrates that DiagCoT (ours) exhibits the highest performance.

\subsubsection{X-Ray Report Generation (RRG)}

The experimental results for the RRG task are presented in Figure~\ref{fig: Result_visual_all} a. This figure demonstrates the comparison of different baseline models, including traditional architectural models and vision-language models. As evidenced by the within-dataset evaluation results in Figure~\ref{fig: Result_visual_all} a (Top), our approach demonstrates statistically significant improvements over multiple baseline methods across most evaluated metrics. In addition, As evidenced by the cross-dataset evaluation results in Figure~\ref{fig: Result_visual_all} a (Bottom), DiagCoT exhibits robust performance on out-of-domain test sets, indicating its effective generalization capability. The slight decrease observed in certain metrics may be attributed to potential overfitting to the in-domain dataset during the third training stage, which could have limited the model's adaptability to external data distributions. Please refer to Section~\ref{sec: Complete Experimental Results Table} for detailed numerical results.

\subsubsection{Fine-tuning classification}
For the downstream classification task, we employed our two-stage trained report generation model, training and evaluating it on the CheXpert~\cite{irvin2019chexpert} dataset, a large-scale public chest radiograph dataset released by the Stanford University School of Medicine in 2019. This dataset likewise consists of chest X-ray images (~210,000 studies) annotated with 14 distinct pathological labels. The task is a multi-label classification problem, meaning a single X-ray image may be associated with multiple pathology labels, such as Cardiomegaly, Lung Opacity, Edema, etc. Images showing no signs of disease are labeled as “No Finding”. Model performance was evaluated using the Curve (AUC) score metrics. The detailed evaluation results are presented in Figure~\ref{fig: Result_visual_all} b.

DiagCoT achieves a mean AUC of 62.8\% across 14 pathologies in the CheXpert multi-label classification task. The model demonstrates notable advantages in specific diagnostic categories: its performance in Enlarged Cardiomediastinum (Enl.) reaches 68.5\% – significantly exceeding all comparators by at least 9.8 percentage points (Qwen2.5-VL-72B-AWQ: 58.7\%). For critical pulmonary conditions, DiagCoT attains 71.0\% in Edema (Ede.)(outperforming ChestX-Reasoner-7B's 69.7\%) and 65.1\% in Atelectasis (Ate.)(surpassing CheXagent-2-3b's 58.5\%). While showing competitive results in No Finding (Nofi.) classification (77.3\%, second to CheXagent-2-3b's 81.2\%), the model maintains performance parity with domain specialists in aggregate metrics (CheXagent-2-3b: 62.7\%). Persistent challenges are observed across all models for Fracture (Fra.) detection (DiagCoT: 58.1\%, highest among all). 
%Please refer to Section~\ref{sec: Complete Experimental Results Table} for detailed numerical results.

% \begin{figure*}[!t]
% \begin{center}
% \includegraphics[width=\textwidth]{figures/Result/Result_visual_2_3.pdf}
% \end{center}

% \caption{\textbf{X-Ray Diease classification performanc evaluation on the CheXpert dataset (left) and  X-Ray Diease pathology grounding evaluation on the RSNA dataset (right).} The left figure displays a heatmap comparing the classification performance of DiagCoT against various baseline models on the CheXpert dataset. Darker colors indicate better classification performance. The right figure demonstrates the grounding capability of DiagCoT compared with baseline models on the RSNA dataset.}
%    \label{fig: Result_visual_2}
%   % \vspace{-.15in}
% \end{figure*}

\subsubsection{Fine-tuning grounding}
% \begin{figure*}[!t]
% \begin{center}
% \includegraphics[width=\textwidth]{figures/Result/Result_visual_3.pdf}
% \end{center}

% \caption{\textbf{X-Ray Diease pathology grounding evaluation on the RSNA dataset.}}
%    \label{fig: Result_visual_3}
%   % \vspace{-.15in}
% \end{figure*}

For the grounding task,we likewise leveraged our two-stage trained report generation model and utilized the RSNA Pneumonia Detection Challenge dataset~\cite{wang2017chestx,shih2019augmenting} (Radiological Society of North America, 2018). This dataset comprises approximately 26,684 frontal chest X-ray images with radiologist-annotated bounding boxes identifying pneumonia regions. However, significant annotation imbalance exists: only 6,012 images (~21\%) contain bounding boxes (positive for pneumonia), while 20,672 images (~79\%) lack bounding boxes (negative for pneumonia). To mitigate this data bias, we implemented a data augmentation strategy focused on the bounding-box-positive samples, applying random rotations, translations, scaling, and horizontal flips to increase their representation. Subsequently, the model was trained and evaluated on this augmented dataset. Model performance was evaluated using the mean Intersection over Union (mIoU) metric, which quantifies the spatial overlap accuracy between predicted bounding boxes and ground-truth annotations. Detailed results are presented in Figure~\ref{fig: Result_visual_all} c.

DiagCoT significantly outperforms all comparative models on the RSNA pneumonia detection task, achieving 0.7307 accuracy and 0.3073 mIoU – representing 103\% and 105\% improvements respectively over the strongest baseline (Qwen2.5-VL-72B-AWQ: 0.3604 ACC, 0.1497 mIoU). This demonstrates exceptional capability in localizing pneumonia manifestations from chest X-rays.

\begin{figure*}[!h]
\begin{center}
\includegraphics[width=\textwidth]{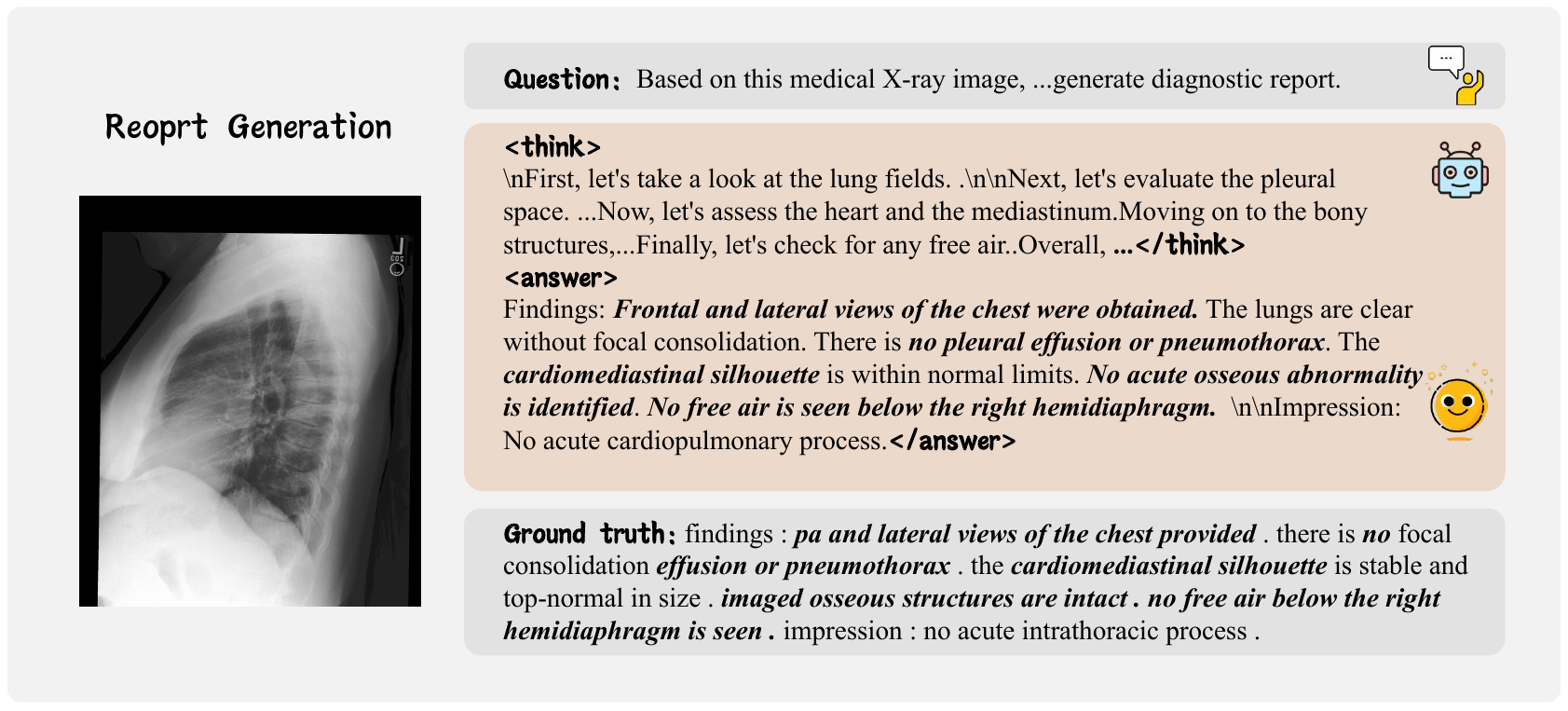}
\end{center}

\caption{\textbf{Qualitative results of Report Generation.}}
   \label{fig: MedLLM-qualitative_1}
  % \vspace{-.15in}
\end{figure*}

\subsection{Qualitative results}

\begin{figure*}[!h]
\begin{center}
\includegraphics[width=\textwidth]{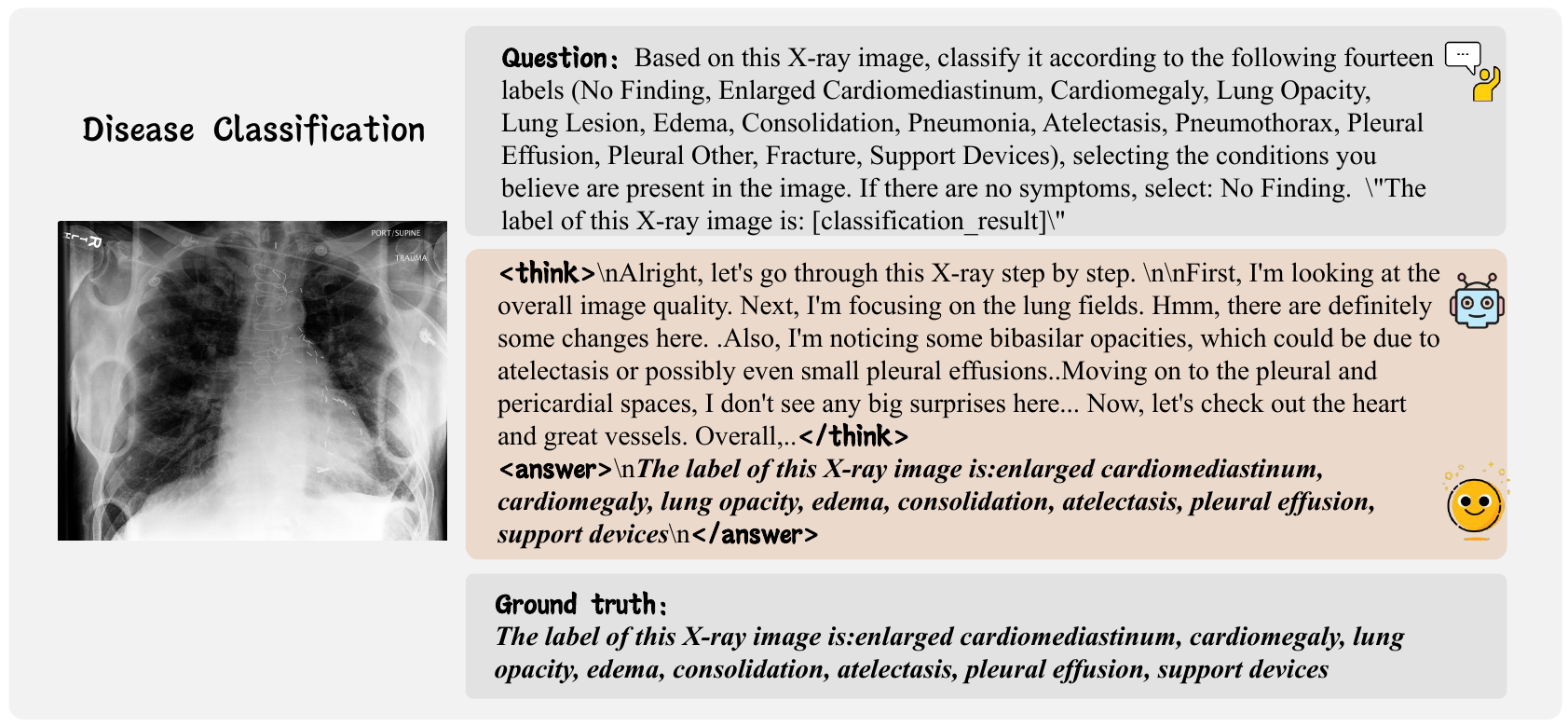}
\end{center}

\caption{\textbf{Qualitative results of Disease Classification.}}
   \label{fig: MedLLM-qualitative_2}
  % \vspace{-.15in}
\end{figure*}

\begin{figure*}[!h]
\begin{center}
\includegraphics[width=\textwidth]{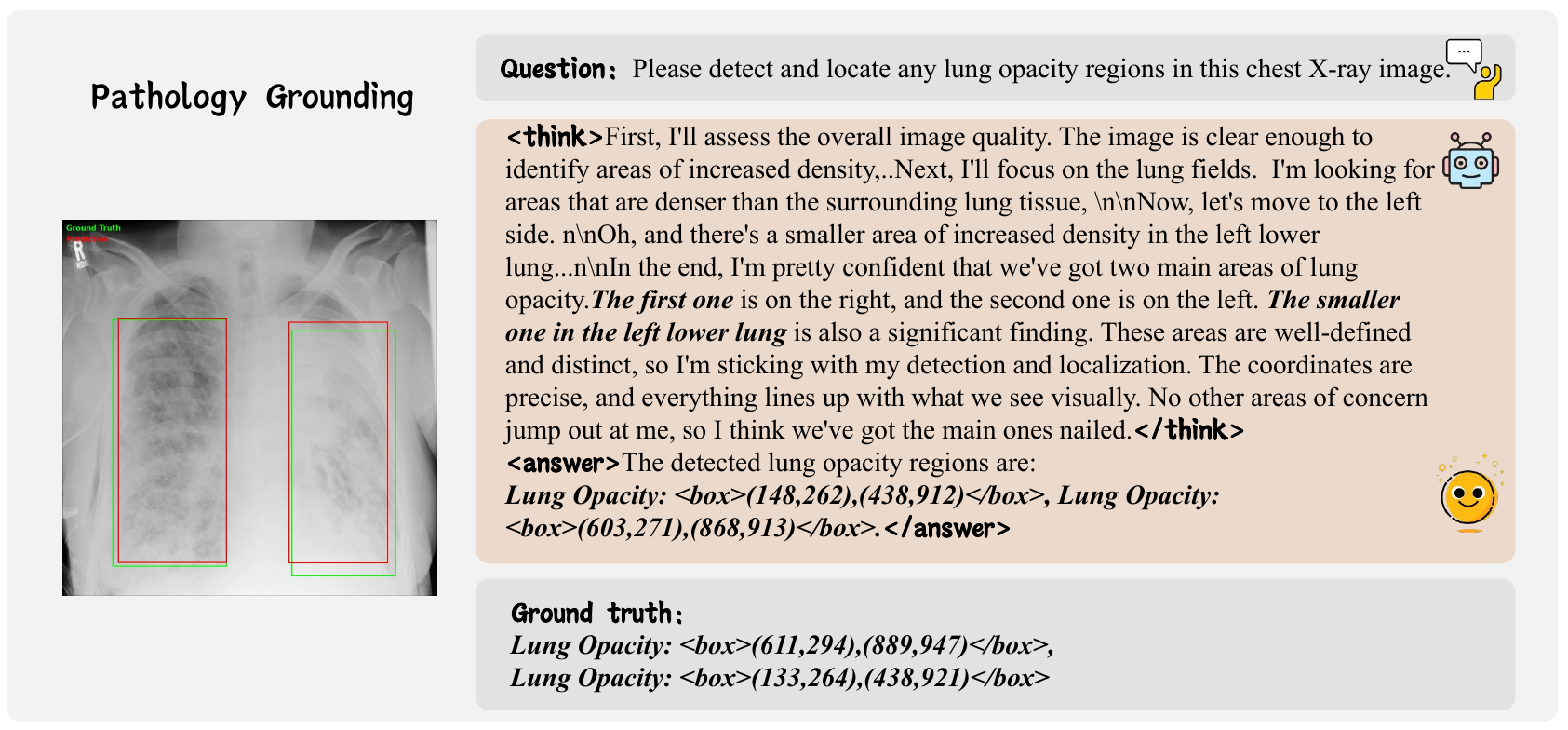}
\end{center}

\caption{\textbf{Qualitative results of Pathology Grounding.}}
   \label{fig: MedLLM-qualitative_3}
  % \vspace{-.15in}
\end{figure*}

% \begin{figure*}[h]
% \begin{center}
% \includegraphics[width=\textwidth]{figures/Method/MedLLM-qualitative-cat.pdf}
% \end{center}

% \caption{\textbf{Qualitative results of Disease classification and Pathology grounding.}}
%    \label{fig: MedLLM-qualitative-cat_2}
%   % \vspace{-.15in}
% \end{figure*}

% \subsection{Ablation Studies}

In this section, we present a qualitative analysis of the report generation, disease classification, and radiology grounding tasks. Visualizations are provided to offer a more intuitive demonstration of DiagCoT's capabilities. Text that is bolded and italicized indicates where the model's prediction aligns with the ground truth label. For a detailed reasoning process, please refer to Section~\ref{sec: Complete Prediction Results Visualization}.

First, for the report generation task, Figure~\ref{fig: MedLLM-qualitative_1} shows that DiagCoT reasoned logically and step-by-step based on the prompts before generating the final diagnostic report. The report demonstrates substantial alignment with the reference (ground truth) report.

Second, Referring to Figure~\ref{fig: MedLLM-qualitative_2}, for the disease classification task, the prompt includes the dataset's label set, presenting the task in a multiple-choice format. Similarly, DiagCoT undergoes a reasoning process before outputting its determined classification label.

Finally, Referring to Figure~\ref{fig: MedLLM-qualitative_3}, for the radiology annotation task, DiagCoT analyzes the image in a logical sequence of comparisons to identify regions of interest. It outputs predicted bounding boxes for abnormal regions (red boxes), while green boxes denote the ground truth annotations. The results indicate that the predicted boxes closely approximate the ground truth boxes in location.

\subsection{Ablation on Training Strategy}

The DiagCoT training strategy comprises three distinct stages: Supervised Fine-Tuning (SFT), SFT incorporating Chain-of-Thought (CoT) data, and Reinforcement Fine-Tuning (RFT). To rigorously evaluate the effectiveness of this multi-stage training regimen, we conduct the following ablation studies:
\begin{itemize}
\item \textbf{DiagCoT-SFT(CoT)}: Train the model directlyusing CoT data, bypassingthe initial SFT Stage (Stage 1).
\item \textbf{DiagCoT-RL(O)}: Only apply RL fine-tuning directly, bypassingboth the initial SFT Stage (Stage 1) and the SFT(CoT) Stage (Stage 2).
\item \textbf{DiagCoT-RL(W.1)}: Incorporate the initial SFT Stage (Stage 1), then proceed directlyto RL fine-tuning, bypassingthe SFT(CoT) Stage (Stage 2).
\item \textbf{DiagCoT-RL(W.2)}: Incorporate the SFT(CoT) Stage (Stage 2), followed by RL fine-tuning, bypassingthe initial SFT Stage (Stage 1).
\item \textbf{DiagCoT}: Our final model, After sequentially progressing through SFT, SFT(CoT), and RL. See detail in Section~\ref{sec3:method}.

\end{itemize}

% \begin{table}[h]
% \centering
% \caption{Ablation on Training Strategy}
% \setlength{\tabcolsep}{8pt} 
% \begin{tabularx}{\textwidth}{lccccccc}
% \toprule
% \text{Model} & \text{BLEU1} & \text{BLEU2} & \text{BLEU3} & \text{BLEU4} & \text{ROUGE-L} & \text{METEOR} & \text{CIDEr} \\
% \midrule
% \text{DiagCoT-SFT(CoT)} & 0.2096 & 
% 0.1130 & 0.0651 & 0.0390 & 0.2184 & 0.0653 & 0.1189\\
% \text{DiagCoT-RL(O)} & 0.2652 & 
% 0.1110 & 0.0488 & 0.0229 & 0.1927 & 0.0262 & 0.1160\\
% \text{DiagCoT-RL(W.1)} & 0.2818 & 
% 0.1697 & 0.1071 & 0.0652 & 0.2395 & 0.0559 & 0.1433 \\
% \text{DiagCoT-RL(W.2)} & 0.3128 & 
% 0.1900 & 0.1243 & 0.0853 &  0.2661 & 0.1118 & \textbf{0.1553}\\

% \text{DiagCoT} & \textbf{0.3260} & 
% \textbf{0.2024} & \textbf{0.1343} & \textbf{0.0900} & \textbf{0.2717} & \textbf{0.1619} & 0.1152\\
% \bottomrule
% \label{tab:Ablation1}
% \end{tabularx}
% \end{table}

\begin{figure*}[!ht]
\begin{center}
\includegraphics[width=\textwidth]{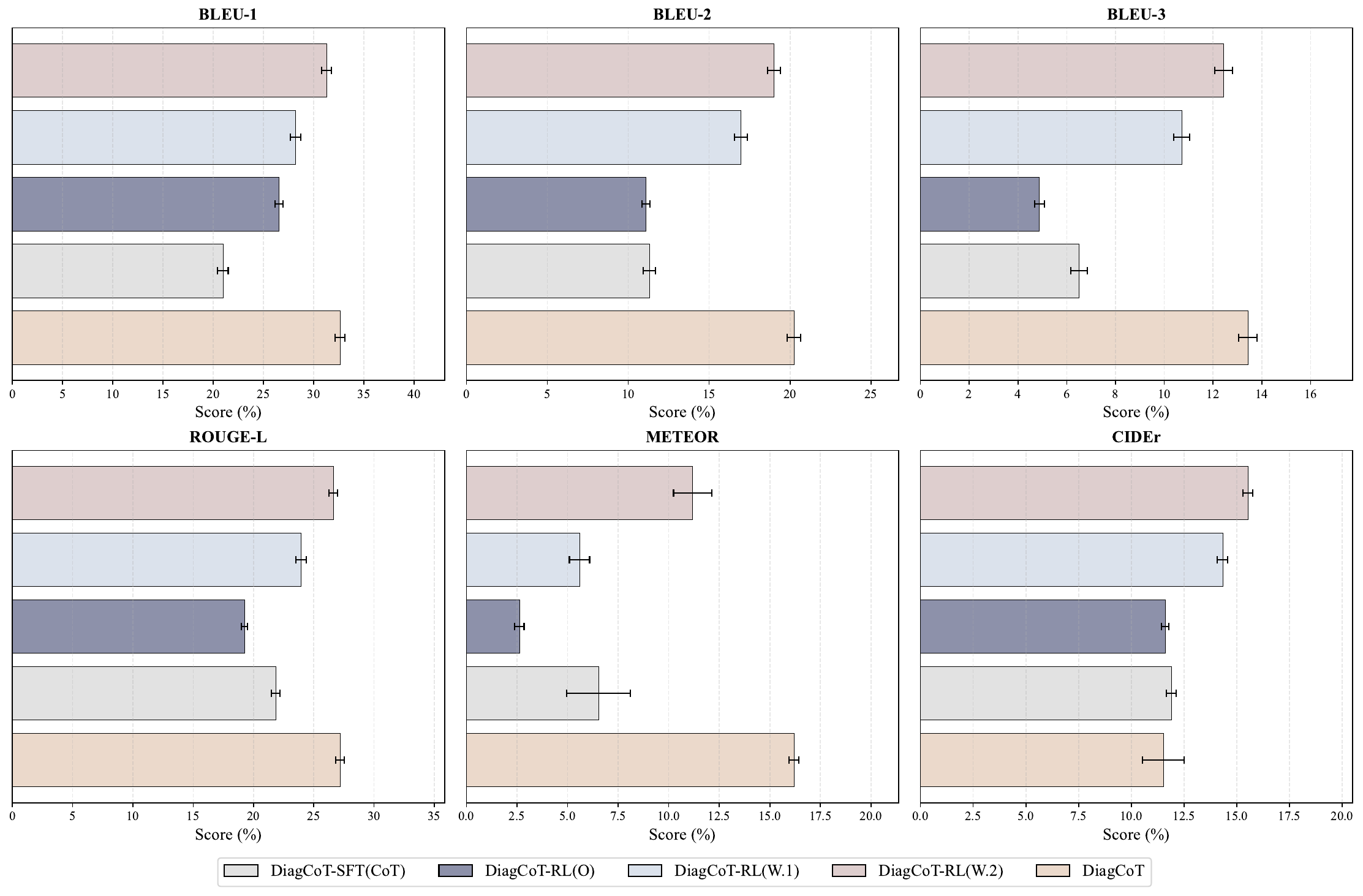}
\end{center}

\caption{\textbf{Ablation Study on Training Strategy.} This figure presents the ablation study results of DiagCoT, primarily validating the feasibility of our complete three-stage pipeline (SFT → SFT(CoT) → RFT). As demonstrated in the figure, four distinct experimental configurations were implemented: DiagCoT-SFT(CoT), DiagCoT-RL(O), DiagCoT-RL(W.1), and DiagCoT-RL(W.2). The evaluation outcomes across multiple metrics indicate that the three-stage training paradigm contributes measurable performance gains to the model.}
   \label{fig: ablation_study_percentage}
  % \vspace{-.15in}
\end{figure*}

As delineated in Figure~\ref{fig: ablation_study_percentage}, our ablation study systematically validates the necessity of each training stage in the proposed DiagCoT training strategy. The complete three-stage pipeline (SFT → SFT(CoT) → RFT) achieves optimal performance on core metrics including BLEU1-4 (0.3260/0.2024/0.1343/0.0900), ROUGE-L (0.2717), and METEOR (0.1619), significantly outperforming all partial variants. Key observations reveal:

(1) Omitting SFT(CoT) (i.e., DiagCoT-RL(W.1: SFT→RFT)) causes severe degradation (-2.72\% BLEU3, -10.60\% METEOR);

(2) Skipping initial SFT (i.e., DiagCoT-RL(W.2: SFT(CoT)→RFT)) reduces BLEU3 by 1\% and fails to match full-pipeline stability;

(3) Direct RL-only training (DiagCoT-RL(O)) collapses completely (BLEU3=0.0488), exhibiting a catastrophic 8.55\% decline in BLEU3 compared to the full model.

Notably, while CIDEr scores marginally favor simplified variants (e.g., 0.1553 for W.2 vs. 0.1152 for DiagCoT), this aligns with our reward design prioritizing accuracy over lexical diversity. These results conclusively demonstrate that the progressive knowledge distillation from factual grounding (SFT) to reasoning capability (SFT(CoT)) and preference alignment (RFT) is the critical success factor. Please refer to Section~\ref{sec: Complete Experimental Results Table} for detailed numerical results.

\section{Discussion}
In this study, we present DiagCoT, a multi-stage diagnostic reasoning framework that significantly enhances vision-language models for chest X-ray interpretation. Our approach uniquely integrates medical knowledge infusion, physician-like reasoning simulation, and reinforcement-based refinement to bridge critical gaps in clinical AI applications. The experimental results across three key tasks—report generation, disease classification, and pathology grounding—demonstrate DiagCoT’s consistent superiority over  baselines.
DiagCoT demonstrates consistent improvements across three diagnostic tasks. (i) For X-ray report generation, it achieves better performance in both within-dataset and cross-dataset evaluations by explicitly modeling clinical reasoning hierarchies. (ii) In disease classification on CheXpert, the framework shows enhanced capability for underrepresented conditions like Enlarged Cardiomediastinum, Edema, and Atelectasis, suggesting improved handling of data-scarce pathologies. (iii) For pathology grounding, DiagCoT achieves higher localization accuracy compared to existing approaches, supporting our hypothesis that structured reasoning chains improve spatial-textual alignment in medical imaging analysis. These results collectively indicate that DiagCoT’s staged training approach effectively enhances diagnostic capabilities for chest X-ray interpretation.

DiagCoT systematically enhances chest X-ray diagnosis through a three-stage training framework (medical knowledge infusion, physician diagnostic thinking simulation, and reasoning process refinement). Experiments demonstrate consistent improvements across report generation, disease classification, and pathology grounding tasks. The structured reasoning chain design effectively strengthens the model’s alignment of medical semantics and spatial features, offering a new paradigm for clinically adaptive multimodal medical AI.

\section{Limitations}
While DiagCoT demonstrates promising results, several limitations warrant attention. First, our study focuses exclusively on chest X-ray interpretation due to public dataset accessibility and computational constraints. We emphasize that the framework is inherently modality-agnostic; its extension to other imaging domains with structured reports (e.g., CT, MRI, ultrasound) represents a critical future direction for validating broader clinical applicability.

Second, DiagCoT builds upon the general-domain VLM Qwen2-VL-7B as its base architecture. Though this model provides a robust foundation, exploring medically pretrained VLMs (e.g., models initialized on radiology-specific corpora) or scaling to larger parameter sizes could further enhance diagnostic precision.

Third, our reinforcement learning stage currently relies on rule-based reward functions. While effective in aligning outputs with clinical logic, future work should develop learnable, domain-specific reward models—trained on expert-annotated reasoning traces—to provide more nuanced, context-aware feedback during optimization.

Lastly, DiagCoT’s evaluation is confined to retrospective datasets. Prospective clinical validation—assessing its impact on radiologist workflow efficiency, diagnostic confidence, and patient management decisions—remains essential prior to real-world deployment.
% retrospective dataset 回顾性数据集 强调整个研究的数据来源是历史病例。这一限制的本质是模型尚未在真实临床工作流中验证，这是医学AI研究常见的局限性表述。
% \clearpage
\section{Methods}
\label{sec3:method}
In this section, we introduce DiagCoT, which is a CoT-guided framework for X-ray Report Generation. Section~\ref{sec3.1:sft} and Section~\ref{sec: Stage 1-Alignment Stage} first elaborates on the conventional SFT training method without reasoning mechanisms, where training without thought modeling refers to direct supervised fine-tuning using original X-ray image-report pairs. This approach is common and efficient for large-scale models, yet for medical report generation, directly using concise original reports fails to capture critical information. We require the model to emulate radiologists in diagnosing X-ray images from multiple perspectives (e.g., image quality, anatomical structures, abnormal radiological findings, and indications of potential diseases) before generating diagnostic reports. Thus, Section~\ref{sec: CoT-Tuning Stage} details the construction of a high-quality CoT dataset, subsequently integrating CoT data with SFT and RFT training to equip the model with domain-specific knowledge for medical imaging reporting.

\begin{figure*}[!htbp]
\begin{center}
\includegraphics[width=\textwidth]{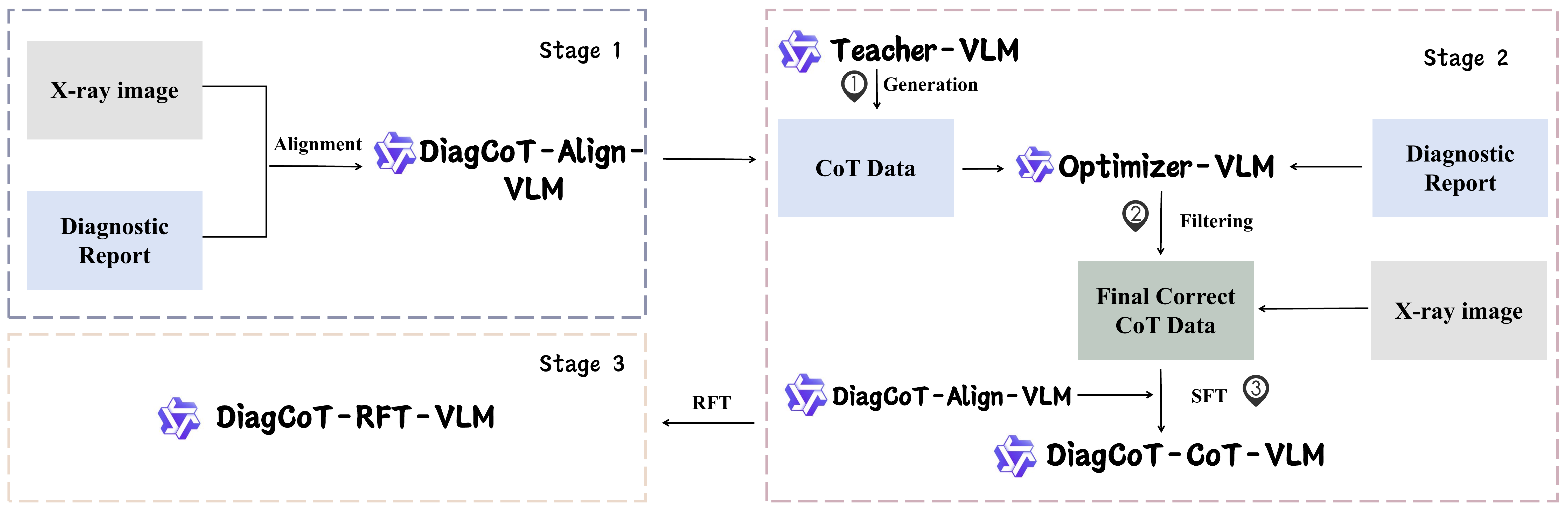}
\end{center}

\caption{\textbf{Overview of the Methods Section}. This figure illustrates the overall training stages of theDiagCoT framework and the models employed. The process comprises three distinct stages: the Alignment Stage, the CoT-Tuning Stage, and the RFT (Reinforcement Fine-Tuning) Stage. The models utilized include a Teacher-VLM (Qwen2.5-VL-32B-LoRA) and an Optimizer-VLM (Qwen2.5-VL-72B-AWQ).}
\label{fig: overview of method}
  % \vspace{-.15in}
\end{figure*}

%方法大图
% \begin{figure*}[!htbp]
% \begin{center}
% \includegraphics[width=\textwidth]{figures/Method/Method_cat_all_v3.pdf}
% \end{center}

% \caption{\textbf{Overview of the Methods Section}. (a) This figure illustrates the overall training stages of theDiagCoT framework and the models employed. The process comprises three distinct stages: the Alignment Stage, the CoT-Tuning Stage, and the RFT (Reinforcement Fine-Tuning) Stage. The models utilized include a Teacher-VLM (Qwen2.5-VL-32B-LoRA) and an Optimizer-VLM (Qwen2.5-VL-72B-AWQ). (b) This figure presents the detailed training pipeline for the Alignment Stage. During this Stage, DiagCoT primarily uses the MIMIC-CXR training set to train the Merger layer, aiming to equip the model with the foundational capability for preliminary medical report generation based on images. (c) This figure depicts the data collection workflow for the Chain-of-Thought (CoT) data during the CoT-Tuning Stage. This collection process employs specific prompting strategies to guide the model through iterative reflection and error correction, resulting in the generation of higher-quality CoT data. (d) This figure shows the filtering process for the CoT dataset. The Optimizer-VLM (Qwen2.5-VL-72B-AWQ) is used to select high-quality reasoning chains that are consistent with the final diagnoses, resulting in a refined training dataset.
% }
% \label{fig: overview of method}
%   % \vspace{-.15in}
% \end{figure*}

\subsection{Preliminaries}
\label{sec3.1:sft}
In the domain of medical imaging report generation, the conventional supervised fine-tuning (SFT) stage utilizes paired medical images $X$ and corresponding manually authored reports $Y=\{y_1,y_2,...,y_K\}$ to perform domain adaptation for the baseline model $F_\theta$. By aligning visual features with radiological language descriptions, this process optimizes the model to generate structured reports, equipping it with the capability to produce clinically compliant reports. Equation~\ref{eq:SFT} presents the auto-regressive conditional generation task, maximizing the log-likelihood of the ground-truth report sequence. The loss function is defined as:
\begin{equation}
\mathcal{L}_{\mathrm{SFT}}=-\sum_{t=1}^T\log P(y_t|y_{<t},X;\theta)
\label{eq:SFT}
\end{equation}
where $X$ denotes the input medical image and $\theta$ represents the trainable parameter set during fine-tuning. Through Equation~\ref{eq:SFT}, the baseline model acquires foundational domain knowledge. Although SFT is straightforward and efficient, conventional supervised fine-tuning methods exhibit critical limitations in medical report generation tasks: implicit reasoning opacity, high risk of autoregressive error propagation, and insufficient structural compliance of reports. To address this, we introduce a Chain-of-Thought (CoT)-enhanced SFT method combined with reinforcement learning to further improve generation quality. This approach explicitly models the multi-stage reasoning process of "visual feature localization → pathological sign inference → diagnostic conclusion generation", while jointly optimizing the generation likelihood of intermediate reasoning chains and final reports.

% \subsection{Training Procedure}
% \label{sec3.2:Traing pro}
% \begin{figure*}[h]
% \begin{center}
% \includegraphics[width=\textwidth]{figures/Method/overall_procedure.pdf}
% \end{center}

% \caption{\textbf{Training Procedure}}
%    \label{fig: Training Procedure}
%   % \vspace{-.15in}
% \end{figure*}

As illustrated in Figure \ref{fig: overview of method}, our method adopts a three-stage progressive training framework:
\begin{enumerate}[label=(\arabic*)]
    \item \textbf{Alignment Stage (Stage 1)}: Aligns the VLM using MIMIC-CXR datasets (image–caption pairs) to establish cross-modal associations between medical images and foundational textual descriptions.
    \item \textbf{CoT-tuning Stage (Stage 2)}: First constructs a CoT dataset, Subsequently, the data was strictly filtered using the Optimizer-VLM to yield a high-quality CoT dataset. Then uses Image–CoT pairs to conduct SFT. This enables the model to learn radiological diagnostic logic and generate structured intermediate reasoning steps.
    \item \textbf{RFT-tuning Stage (Stage 3)}: Optimizes report generation quality via reinforcement learning, producing the final Vision-Language Model (DiagCoT) that ensures diagnostic accuracy and report compliance.
\end{enumerate}
% Details are provided in subsequent sections.

% \begin{figure*}[h]
% \begin{center}
% \includegraphics[width=\textwidth]{figures/Method/merger_new.pdf}
% \end{center}

% \caption{\textbf{Alignment Stage}}
%    \label{fig: Alignment Stage}
%   % \vspace{-.15in}
% \end{figure*}

\begin{figure*}[!htbp]
\begin{center}
\includegraphics[width=\textwidth]{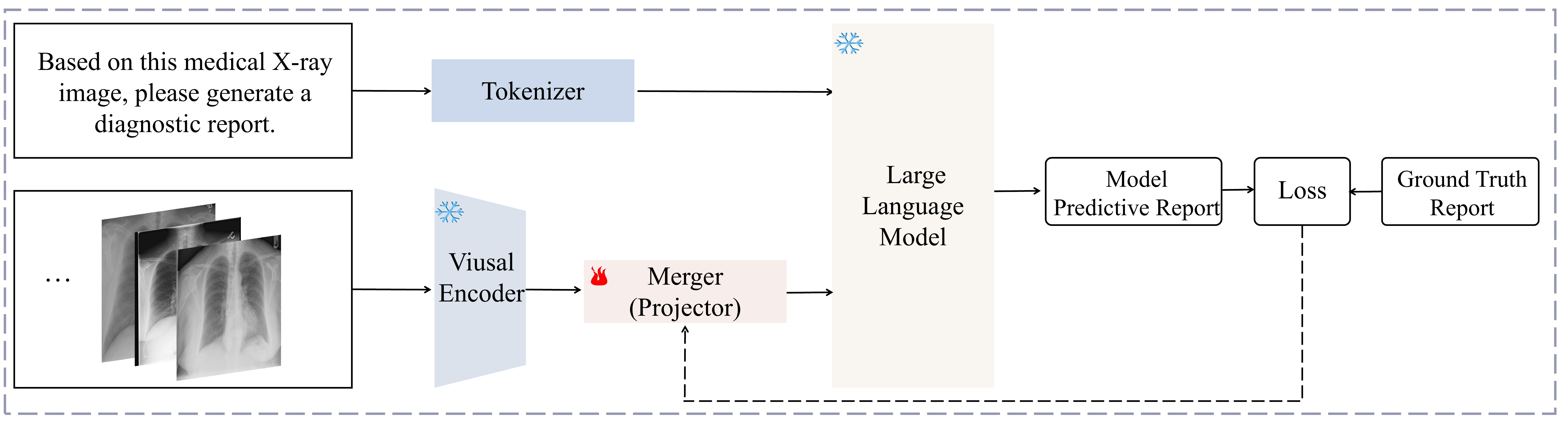}
\end{center}

\caption{\textbf{Stage 1: Merger training.} This figure presents the detailed training pipeline for the Alignment Stage. During this Stage, DiagCoT primarily uses the MIMIC-CXR training set to train the Merger layer, aiming to equip the model with the foundational capability for preliminary medical report generation based on images.}
\label{fig: Stage1-Merger training}
  % \vspace{-.15in}
\end{figure*}

\subsection{Stage 1: Alignment Stage}
\label{sec: Stage 1-Alignment Stage}
Given the inherent limitations of baseline models in the generation of medical reports, we adopt alignment operations from LLaVA-Med~\cite{li2023llava} and MoColl~\cite{yang2025mocoll} to align medical images with reports for the baseline model, thereby establishing fundamental report generation capabilities. As detailed in Figure \ref{fig: Stage1-Merger training}: First, preprocess the original MIMIC-CXR dataset to obtain image-report pairs, with prompts standardized as: "Based on this medical X-ray image, please generate a diagnostic report." Second, feed both text and images into the baseline model to obtain predicted reports; these are then compared against ground-truth reports for updating via the autoregressive loss in Equation~\ref{eq:SFT}. During updating, we keep the LLM and Visual Encoder frozen, updating only the Merger (Projector) layer. This ultimately results in the report-aligned model $\mathcal{M}_\text{Stage1}$.

\subsection{Stage 2: CoT-Tuning Stage}
\label{sec: CoT-Tuning Stage}
The CoT fine-tuning stage comprises three sub-stages: Expert model construction; CoT data collection and Filtering; and CoT fine-tuning. \\
\textbf{(1) Expert Model Construction.}
To generate high-quality CoT data, we require a model proficient in medical report generation. Thus, we construct an expert model endowed with domain-specific knowledge. Considering deployment and computational constraints, we adopt Qwen2.5-VL-32B-LoRA as the Teacher-VLM. Using the MIMIC-CXR dataset, we perform fine-tuning via LoRA (Low-Rank Adaptation) to develop an expert model capable of fundamental medical report generation $\mathcal{M}_\text{med init}$. \\

\begin{figure*}[h]
\begin{center}
\includegraphics[width=\textwidth]{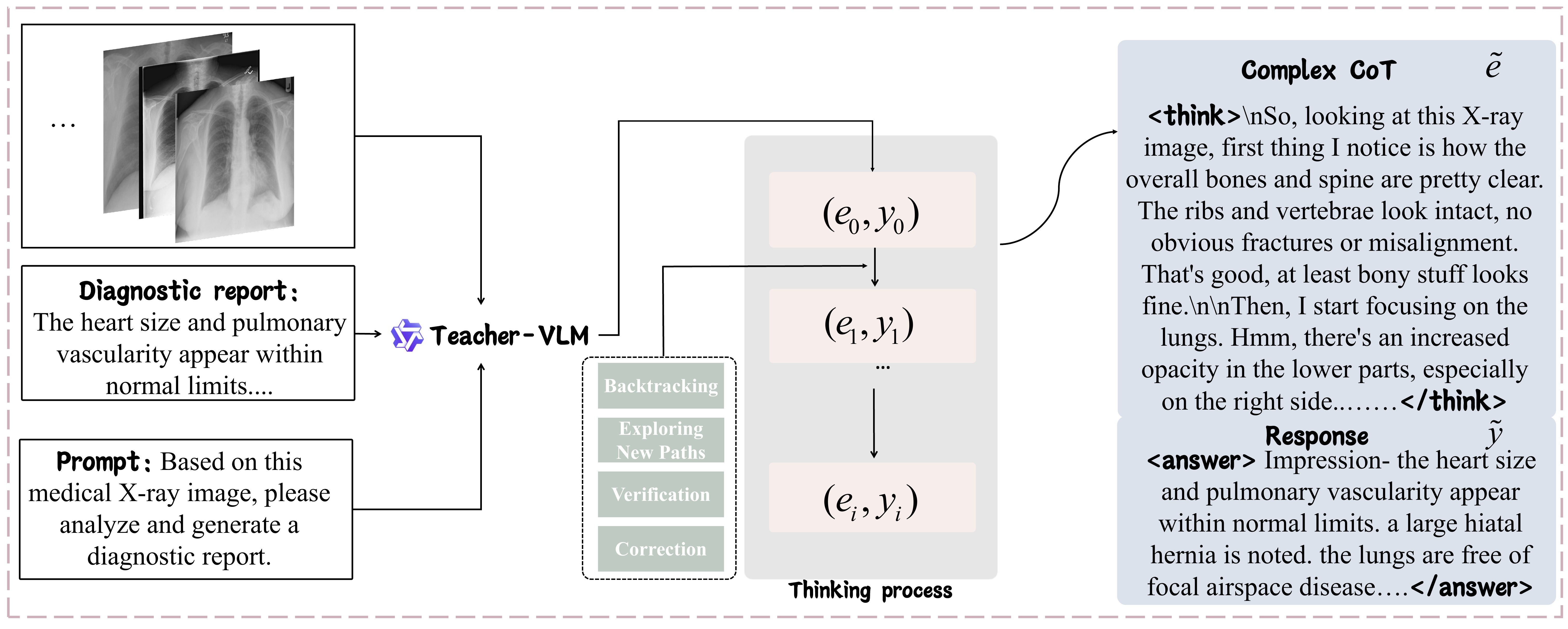}
\end{center}

\caption{\textbf{Stage 2: CoT Collection.} This figure depicts the data collection workflow for the Chain-of-Thought (CoT) data during the CoT-Tuning Stage. This collection process employs specific prompting strategies to guide the model through iterative reflection and error correction, resulting in the generation of higher-quality CoT data.}
    \label{fig: CoT Collection}
  % \vspace{-.15in}
\end{figure*}

\textbf{(2) CoT Data Collection and Filtering.}
As shown in Figure \ref{fig: CoT Collection}, Teacher-VLM serves as the expert model obtained from the first substage. We adapt HuatuoGPT-o1's~\cite{chen2024huatuogpto1medicalcomplexreasoning} methodology for generating chains of thought to the vision-language domain, specifically targeting X-ray report generation. For details of the prompt design, refer to Section~\ref{sec: CoT Collection Prompt Design}. First, as shown on the left of Figure~\ref{fig: CoT Collection}, the inputs consist of three parts: X-ray image $X=\{x_1,x_2,...,x_k\}$, prompt $P=\{p_1,p_2,...,p_k\}$ , and reference report $R=\{r_1,r_2,...,r_k\}$. The prompt and reference report are concatenated as textual input $Y=\{({{x}_{1}},{{r}_{1}}),({{x}_{2}},{{r}_{2}}),...,({{x}_{k}},{{r}_{k}})\}$. After receiving the corresponding image and text, the model performs initialization:
\begin{equation}
({{e}_{0}},{{y}_{0}})={{\mathcal{M}}_{\text{med init}}}(X,Y);
\end{equation}
Here MLLM refers to the trained Teacher-VLM, denotes image and text inputs (X-ray image and corresponding prompt/reference report). $e_0$ and $y_0$represent the initial reasoning process and generated report after initialization. Subsequently, we validate the model's initial output $e_0$ and $y_0$. If the verified answer is incorrect, one of four ${{c}_{i}}\in C$ strategies is randomly selected to generate new reasoning steps and answers:
\begin{equation}
    ({{e}_{i}},{{y}_{i}})=\mathcal{M}_{\text{med init}}^{{{c}_{i}}}(x,[{{e}_{0}},{{y}_{0}},\ldots ,{{e}_{i-1}},{{y}_{i-1}}])
\end{equation}

\begin{figure*}[!htb]
\begin{center}
\includegraphics[width=\textwidth]{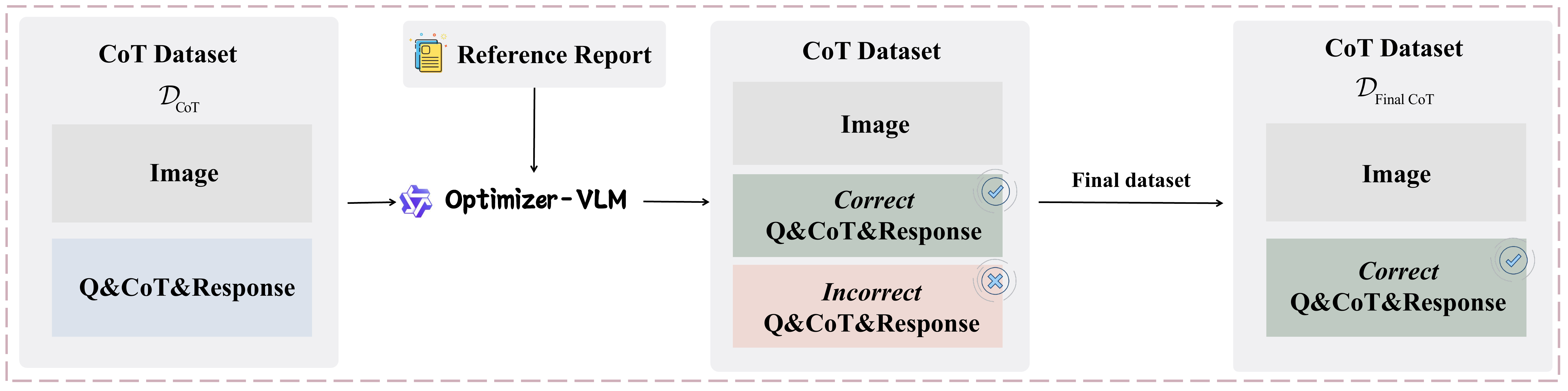}
\end{center}

\caption{\textbf{Stage 2: CoT Data Filtering.} This figure shows the filtering process for the CoT dataset. The Optimizer-VLM (Qwen2.5-VL-72B-AWQ) is used to select high-quality reasoning chains that are consistent with the final diagnoses, resulting in a refined training dataset.}
    \label{fig: CoT extract}
  % \vspace{-.15in}
\end{figure*}

Here $i$ denotes the response generated at the $i$-th iteration. We define four strategies for model reflection and correction:\\
\textbf{Explore new pathways}: MLLM attempts a new approach $e_i$ differing from prior reasoning path $\{{{e}_{0}},...,{{e}_{i-1}}\}$, generating corresponding answer $y_i$. This process emphasizes innovative diversity in reasoning paths.\\
\textbf{Backtracking mechanism}: At early reasoning stages $(j<i-1)$, MLLM can backtrack to historical path $\{{{e}_{j}},{{y}_{j}}\}$ to extend reasoning. This provides correction opportunities during early exploration.\\
\textbf{Self-verification mechanism}: MLLM evaluates the completeness of current reasoning chain $\{{{e}_{i-1}},{{y}_{i-1}}\}$ through verification process $e_i$, outputting verified conclusion $y_i$. This stage forms a quality-check loop.\\
\textbf{Critique-correction mechanism}: MLLM conducts logical review of existing reasoning process $\{{{e}_{i-1}},{{y}_{i-1}}\}$, generating revised reasoning path $e_j$ and optimized answer $y_j$. This process incorporates self-correction functionality.

This iteration continues until $y_i$is verified as correct. Considering time costs, we set maximum attempts to 3. If no correct answer is obtained after maximum attempts, $({{x}_{i}},{{y}_{i}})$ is discarded. Upon successfully obtaining a series of reasoning steps and answers $[{{e}_{0}},{{y}_{0}},...,{{e}_{i}},{{y}_{i}}]$, formatting aligns the reasoning process with human cognitive patterns.
\begin{equation}
    \tilde{e},\tilde{y}\leftarrow \mathcal{M}_{\text{med init}}^{\text{Reformat}}([{{e}_{0}},{{y}_{0}},\ldots ,{{e}_{i}},{{y}_{i}}])
\end{equation}

$\tilde{e}$ is the final complex chain of thought obtained, which reflects the model's more meticulous reasoning and observation of X-ray images. 

After the Collection, we obtain dataset $\{{{x}_{1}},{{\tilde{e}}_{1}},{{\tilde{y}}_{1}},...,{{x}_{n}},{{\tilde{e}}_{n}},{{\tilde{y}}_{n}}\}\in {{\mathcal{D}}_{\text{CoT}}}$. Although diverse strategies were employed to guide and correct reasoning chains and responses during this stage, errors may exist in reasoning processes or final outputs. Therefore, we filter to ensure correctness of the CoT dataset ${{\mathcal{D}}_{\text{CoT}}}$. The specific process is shown in Figure \ref{fig: CoT extract}: First, we use a Optimizer-VLM as the expert model, inputting data ${{\mathcal{D}}_{\text{CoT}}}$ from and reference reports into this expert model; Second, the expert model compares them with particular attention to whether CoT reasoning aligns with reference reports; Finally, the expert model categorizes data into consistent and inconsistent sets. We retain consistent data as our final CoT dataset ${{\mathcal{D}}_{\text{Final CoT}}}$.

\textbf{(3) CoT fine-tuning.}
After obtaining the high-quality Final CoT dataset ${{\mathcal{D}}_{\text{Final CoT}}}$, we perform SFT fine-tuning based on the first-stage model $\mathcal{M}_\text{Stage1}$. During this fine-tuning phase, we freeze the Visual Encoder and only train the Merger and the LLM components. This ultimately results in the report-aligned model $\mathcal{M}_\text{Stage2}$.

\subsection{Stage 3: RFT Stage}
Following Stage 2, $\mathcal{M}_\text{Stage2}$ trained with high-quality CoT data can analyze a given X-ray image from multiple perspectives before generating diagnostic reports. To further enhance the accuracy of $\mathcal{M}_\text{Stage2}$ reasoning chains and diagnostic reports, Stage 3 employs the GRPO algorithm for reinforcement learning training—a lightweight alternative to PPO that reduces reward model costs by using rule-based rewards and group-relative comparisons. While GRPO traditionally addresses domains like mathematics and coding (often with multiple-choice formats), our open-ended medical report generation task utilizes custom reward functions. These are categorized into format rewards and precision rewards, detailed later in the reward design section.

\begin{figure*}[!ht]
\begin{center}
\includegraphics[width=\textwidth]{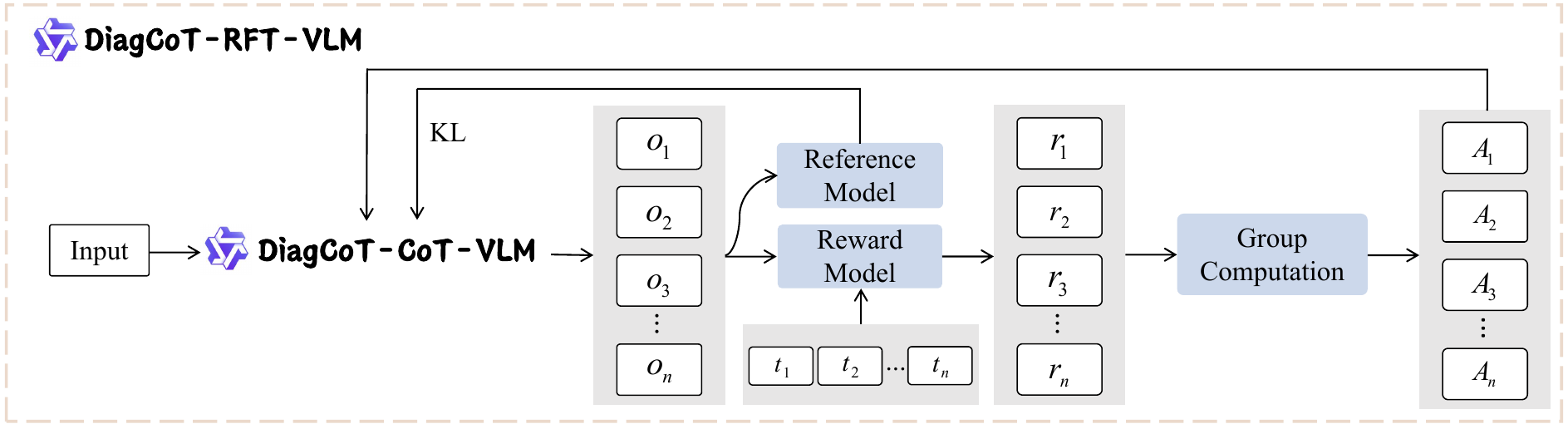}
\end{center}

\caption{\textbf{Stage 3: RFT Stage.} This figure illustrates the overall workflow of the reinforcement learning process in the third stage of DiagCoT. Starting from the left, the input consists of $x$ and $q$, which represent a medical image and a query sampled from the RFT dataset, respectively. These are fed into the DiagCoT-COT model trained in the second stage $\mathcal{M}_\text{Stage2}$, which generates multiple responses, denoted as $o_1, o_2, \dots, o_n$. Each of these responses is then input into both a reference model and a reward model. The reference model acts as a constraint to prevent the output from deviating excessively from its baseline. Meanwhile, the reward model assigns rule-based scores to each response using ground-truth labels $t_1, t_2, \dots, t_n$, resulting in a specific reward value $r_1, r_2, \dots, r_n$ for each generated response. Finally, the relative advantages $A_1, A_2, \dots, A_n$ within the group are computed to optimize the model, yielding the final DiagCoT-RFT-VLM. }
   \label{fig: RFT}
  % \vspace{-.15in}
\end{figure*}

\textbf{(1) GRPO Loss Construction.}
First, $x,q$ represents a medical image and query sampled from $p({{D}_{\text{RFT}}})$. ${{\pi }_{{{\theta }_{\text{old}}}}}$ denotes the old policy model, the updated policy model ${{\pi }_{{{\theta }_{\text{new}}}}}$ after iterative optimization, and the initial reference model ${{\pi }_{\text{ref}}}$ (kept frozen during deployment). indicates the number of responses per policy for group advantage calculation. The loss function is structured as follows:
% $$
% \begin{matrix}
%    {{\mathcal{J}}_{\text{GRPO}}}(\theta )={{\mathbb{E}}_{x,q\sim P({{D}_{\text{RFT}}}),\{{{o}_{i}}\}_{i=1}^{G}\sim {{\pi }_{{{\theta }_{\text{old}}}}}(O|x,q)}}\frac{1}{G}\underset{i=1}{\overset{G}{\mathop{\sum }}}\,\left[ \min \left( \frac{{{\pi }_{{{\theta }_{\text{new}}}}}({{o}_{i}}\mid x,q)}{{{\pi }_{{{\theta }_{\text{old}}}}}({{o}_{i}}\mid x,q)}{{A}_{i}},clip\left( \frac{{{\pi }_{{{\theta }_{\text{new}}}}}({{o}_{i}}\mid x,q)}{{{\pi }_{{{\theta }_{\text{old}}}}}({{o}_{i}}\mid x,q)},1-\epsilon ,1+\epsilon  \right){{A}_{i}} \right)-\beta {{\text{D}}_{\text{KL}}}\left( \left. {{\pi }_{{{\theta }_{\text{new}}}}} \right\|{{\pi }_{\text{ref}}} \right) \right]  \\
% \end{matrix}
% $$

\begin{equation}
\begin{aligned}
\mathcal{J}_{\text{GRPO}}(\theta) 
= \mathbb{E}_{x, q \sim P(D_{\text{RFT}}),\, \{o_i\}_{i=1}^{G} \sim \pi_{\theta_{\text{old}}}(O|x,q)} \frac{1}{G} 
\sum_{i=1}^{G} \Bigg[ &\min\left( 
\frac{\pi_{\theta_{\text{new}}}(o_i|x,q)}{\pi_{\theta_{\text{old}}}(o_i|x,q)} A_i, \right. \\
&\left.\text{clip}\left( 
\frac{\pi_{\theta_{\text{new}}}(o_i|x,q)}{\pi_{\theta_{\text{old}}}(o_i|x,q)}, 1 - \epsilon, 1 + \epsilon 
\right) A_i \right) \\
& - \beta\, \text{D}_{\text{KL}} \left( 
\pi_{\theta_{\text{new}}} \, \| \, \pi_{\text{ref}} 
\right) \Bigg]
\end{aligned}
\end{equation}

The core term $\min \left( \frac{{{\pi }_{{{\theta }_{\text{new}}}}}({{o}_{i}}\mid x,q)}{{{\pi }_{{{\theta }_{\text{old}}}}}({{o}_{i}}\mid x,q)}{{A}_{i}},\text{clip}\left( \frac{{{\pi }_{{{\theta }_{\text{new}}}}}({{o}_{i}}\mid x,q)}{{{\pi }_{{{\theta }_{\text{old}}}}}({{o}_{i}}\mid x,q)},1-\epsilon ,1+\epsilon  \right){{A}_{i}} \right)$ applies the ratio of old-to-new policy probabilities multiplied by advantage function $A_i$ . A clipping function $\left[ 1-\epsilon ,1+\epsilon  \right]$ constrains policy updates within to prevent instability. Simultaneously, a KL divergence term $\beta {{\text{D}}_{\text{KL}}}\left( \left. {{\pi }_{{{\theta }_{\text{new}}}}} \right\|{{\pi }_{\text{ref}}} \right)$ limits deviation between current policy ${{\pi }_{{{\theta }_{\text{new}}}}}$ and reference policy ${{\pi }_{\text{ref}}}$. Through batch sampling (G groups) and expectation calculation ${{\mathbb{E}}_{x,q\sim P(D)}}$, this enables stable policy optimization balancing robustness and efficiency.

\textbf{(2) Reward Function Design.}
Our medical report generation incorporates two rewards:
Format Reward ${R}_{\text{format}}$: Requires output structured as <think></think><answer></answer>. The  tag <think></think> contains multi-perspective analysis of medical images; <answer></answer> delivers final reports. This activates the model’s intrinsic reasoning capability.
Precision Reward ${R}_{\text{acc}}$: Combines weighted scores of BLEU1-4, ROUGE-L, METEOR, and CIDEr to assess similarity between generated and reference reports.
The composite reward function is:
\begin{equation}
    {{R}_{\text{all}}}={{R}_{\text{format}}}+{{R}_{\text{acc}}}
\end{equation}
\section{Related Work}
\label{sec:headings}

\subsection{General VLMs and Medical VLMs}
Although general vision-language models (e.g., CLIP~\cite{radford2021learning}, Qwen2.5-VL~\cite{bai2023qwenvlversatilevisionlanguagemodel}) excel in natural image understanding, the significant semantic gap between their pretraining objectives and medical imaging substantially limits clinical applications. Particularly in X-ray report generation tasks, these generic models struggle to accurately capture complex relationships between critical imaging findings (e.g., ground-glass opacity, pneumothorax line) and diagnostic conclusions, resulting in reports lacking clinical credibility. Current medical VLMs (e.g., LLaVA-Med~\cite{li2023llava}, CXR-LLAVA~\cite{lee2025cxr}) adapt to radiological data through supervised fine-tuning (SFT), which can generate structured report text but suffer from broken decision chains—SFT’s token-by-token prediction loss fails to authentically model the clinical reasoning process from observation to diagnosis. Notably, paradigm innovations in VLM training offer potential to overcome these limitations. Reinforcement learning from human feedback (RLHF)~\cite{christiano2017deep} techniques, validated in pure text models (LLMs) for regulating complex output spaces, remain underexplored for medical multimodal tasks like X-ray report generation. We propose that integrating SFT with reinforcement learning can drive models to generate radiologic reports combining precision with interpretability.

\subsection{Reinforcement Learning}
Reinforcement learning (RL), as a core technique for optimizing sequential decision-making, achieves remarkable success in high-frequency interaction scenarios like game agent training, robotic control, and resource scheduling. Among RL algorithms, proximal policy optimization (PPO)~\cite{schulman2017proximal} has become the mainstream paradigm in deep reinforcement learning due to its advantage of ensuring training stability through constrained policy updates. However, PPO relies on the joint optimization of policy and value function networks, presenting challenges of high computational costs and implementation complexity in complex tasks.

Consequently, the GRPO~\cite{shao2024deepseekmathpushinglimitsmathematical} algorithm has emerged. By eliminating the value function network in PPO and calculating relative advantages through in-group relative return mechanisms, GRPO streamlines the optimization process. In the medical domain, Med-R1~\cite{lai2025med} enhances medical visual question answering capabilities through GRPO's reward-driven learning that transcends static annotations. For natural images, Visual-RFT~\cite{liu2025visual} employs GRPO with task-specific reward functions (e.g., Intersection-over-Union rewards for object detection). Vision-R1~\cite{huang2025vision} utilizes GRPO to improve mathematical reasoning abilities.

This study introduces GRPO for third-stage optimization, automatically generating reward signals solely based on BLEU-1 to BLEU-4 and ROUGE-L NLG metrics to enhance the model's clinical reasoning and report generation capabilities.

\section{Data availability}
All datasets used in this study are publicly accessible. This study utilized datasets that are all publicly
accessible.

\section{Code availability}
The code supporting the findings of this study will be made publicly available upon publication.

\section{Acknowledgments}
This work was supported in part by Shenzhen Science and Technology Program under grant no. JCYJ20240813155840052; the National Key R$\&$D Program of China (2022YFA1004203, 2021YFF0501503), the National Natural Science Foundation of China (62125111, 62331028, 62476268, 62206273).

\section{Author Contributions}
Author Contributions: All listed authors meet the ICMJE four criteria for authorship. Y.L. and W.H. contributed equally to this work. Z.C. and D.L. are the corresponding authors. All authors (Y.L., W.H., Z.C. and D.L.) contributed to the conception and design of the study. Y.L. and Z.C. drafted the manuscript, while W.H., Z.C. and D.L. provided critical revisions for important intellectual content. All authors approved the final version for publication and agree to be accountable for all aspects of the work, ensuring that questions related to the accuracy or integrity of any part of the work are appropriately investigated and resolved.

\section{Competing Interests}
We declare that the authors have no competing interests as defined by Nature Portfolio, or other interests that might be perceived to influence the results and/or discussion reported in this paper.

\clearpage
\bibliographystyle{unsrt}
\bibliography{main} % Entries are in the refs.bib file

\begin{thebibliography}{10}

\bibitem{liu2023visual}
Haotian Liu, Chunyuan Li, Qingyang Wu, and Yong~Jae Lee.
\newblock Visual instruction tuning.
\newblock {\em Advances in neural information processing systems}, 36:34892--34916, 2023.

\bibitem{bai2023qwenvlversatilevisionlanguagemodel}
Jinze Bai, Shuai Bai, Shusheng Yang, Shijie Wang, Sinan Tan, Peng Wang, Junyang Lin, Chang Zhou, and Jingren Zhou.
\newblock Qwen-vl: A versatile vision-language model for understanding, localization, text reading, and beyond, 2023.

\bibitem{liu2024deepseek}
Aixin Liu, Bei Feng, Bin Wang, Bingxuan Wang, Bo~Liu, Chenggang Zhao, Chengqi Dengr, Chong Ruan, Damai Dai, Daya Guo, et~al.
\newblock Deepseek-v2: A strong, economical, and efficient mixture-of-experts language model.
\newblock {\em arXiv preprint arXiv:2405.04434}, 2024.

\bibitem{wang2024cogvlm}
Weihan Wang, Qingsong Lv, Wenmeng Yu, Wenyi Hong, Ji~Qi, Yan Wang, Junhui Ji, Zhuoyi Yang, Lei Zhao, Song XiXuan, et~al.
\newblock Cogvlm: Visual expert for pretrained language models.
\newblock {\em Advances in Neural Information Processing Systems}, 37:121475--121499, 2024.

\bibitem{monshi2020deep}
Maram Mahmoud~A Monshi, Josiah Poon, and Vera Chung.
\newblock Deep learning in generating radiology reports: A survey.
\newblock {\em Artificial Intelligence in Medicine}, 106:101878, 2020.

\bibitem{2015Deep}
Andrej Karpathy and Li~Fei-Fei.
\newblock Deep visual-semantic alignments for generating image descriptions.
\newblock In {\em Computer Vision \& Pattern Recognition}, 2015.

\bibitem{2015Long}
Jeff Donahue, Lisa~A Hendricks, Sergio Guadarrama, Marcus Rohrbach, Subhashini Venugopalan, Kate Saenko, and Trevor Darrell.
\newblock Long-term recurrent convolutional networks for visual recognition and description.
\newblock {\em Elsevier,}, 2015.

\bibitem{2015Show}
Kelvin Xu, Jimmy Ba, Ryan Kiros, Kyunghyun Cho, Aaron Courville, Ruslan Salakhutdinov, Richard Zemel, and Yoshua Bengio.
\newblock Show, attend and tell: Neural image caption generation with visual attention.
\newblock {\em Computer Science}, pages 2048--2057, 2015.

\bibitem{2018Bottom}
Peter Anderson, Xiaodong He, Chris Buehler, Damien Teney, and Lei Zhang.
\newblock Bottom-up and top-down attention for image captioning and visual question answering.
\newblock In {\em 2018 IEEE/CVF Conference on Computer Vision and Pattern Recognition (CVPR)}, 2018.

\bibitem{wei2022chain}
Jason Wei, Xuezhi Wang, Dale Schuurmans, Maarten Bosma, Fei Xia, Ed~Chi, Quoc~V Le, Denny Zhou, et~al.
\newblock Chain-of-thought prompting elicits reasoning in large language models.
\newblock {\em Advances in neural information processing systems}, 35:24824--24837, 2022.

\bibitem{johnson2019mimic}
Alistair~EW Johnson, Tom~J Pollard, Nathaniel~R Greenbaum, Matthew~P Lungren, Chih-ying Deng, Yifan Peng, Zhiyong Lu, Roger~G Mark, Seth~J Berkowitz, and Steven Horng.
\newblock Mimic-cxr-jpg, a large publicly available database of labeled chest radiographs.
\newblock {\em arXiv preprint arXiv:1901.07042}, 2019.

\bibitem{demner2015preparing}
Dina Demner-Fushman, Marc~D Kohli, Marc~B Rosenman, Sonya~E Shooshan, Laritza Rodriguez, Sameer Antani, George~R Thoma, and Clement~J McDonald.
\newblock Preparing a collection of radiology examinations for distribution and retrieval.
\newblock {\em Journal of the American Medical Informatics Association}, 23(2):304--310, 2015.

\bibitem{papineni2002bleu}
Kishore Papineni, Salim Roukos, Todd Ward, and Wei-Jing Zhu.
\newblock Bleu: a method for automatic evaluation of machine translation.
\newblock In {\em Proceedings of the 40th annual meeting of the Association for Computational Linguistics}, pages 311--318, 2002.

\bibitem{lin2004rouge}
Chin-Yew Lin.
\newblock Rouge: A package for automatic evaluation of summaries.
\newblock In {\em Text summarization branches out}, pages 74--81, 2004.

\bibitem{denkowski2011meteor}
Michael Denkowski and Alon Lavie.
\newblock Meteor 1.3: Automatic metric for reliable optimization and evaluation of machine translation systems.
\newblock In {\em Proceedings of the sixth workshop on statistical machine translation}, pages 85--91, 2011.

\bibitem{vedantam2015cider}
Ramakrishna Vedantam, C~Lawrence~Zitnick, and Devi Parikh.
\newblock Cider: Consensus-based image description evaluation.
\newblock In {\em Proceedings of the IEEE conference on computer vision and pattern recognition}, pages 4566--4575, 2015.

\bibitem{li2023llava}
Chunyuan Li, Cliff Wong, Sheng Zhang, Naoto Usuyama, Haotian Liu, Jianwei Yang, Tristan Naumann, Hoifung Poon, and Jianfeng Gao.
\newblock Llava-med: Training a large language-and-vision assistant for biomedicine in one day.
\newblock {\em Advances in Neural Information Processing Systems}, 36:28541--28564, 2023.

\bibitem{lee2025cxr}
Seowoo Lee, Jiwon Youn, Hyungjin Kim, Mansu Kim, and Soon~Ho Yoon.
\newblock Cxr-llava: a multimodal large language model for interpreting chest x-ray images.
\newblock {\em European Radiology}, pages 1--13, 2025.

\bibitem{fan2025chestx}
Ziqing Fan, Cheng Liang, Chaoyi Wu, Ya~Zhang, Yanfeng Wang, and Weidi Xie.
\newblock Chestx-reasoner: Advancing radiology foundation models with reasoning through step-by-step verification.
\newblock {\em arXiv preprint arXiv:2504.20930}, 2025.

\bibitem{chen2024chexagent}
Zhihong Chen, Maya Varma, Jean-Benoit Delbrouck, Magdalini Paschali, Louis Blankemeier, Dave Van~Veen, Jeya Maria~Jose Valanarasu, Alaa Youssef, Joseph~Paul Cohen, Eduardo~Pontes Reis, et~al.
\newblock Chexagent: Towards a foundation model for chest x-ray interpretation.
\newblock {\em arXiv preprint arXiv:2401.12208}, 2024.

\bibitem{lu2024deepseek}
Haoyu Lu, Wen Liu, Bo~Zhang, Bingxuan Wang, Kai Dong, Bo~Liu, Jingxiang Sun, Tongzheng Ren, Zhuoshu Li, Hao Yang, et~al.
\newblock Deepseek-vl: towards real-world vision-language understanding.
\newblock {\em arXiv preprint arXiv:2403.05525}, 2024.

\bibitem{chen-emnlp-2020-r2gen}
Zhihong Chen, Yan Song, Tsung-Hui Chang, and Xiang Wan.
\newblock Generating radiology reports via memory-driven transformer.
\newblock In {\em Proceedings of the 2020 Conference on Empirical Methods in Natural Language Processing}, November 2020.

\bibitem{chen-acl-2021-r2gencmn}
Zhihong Chen, Yaling Shen, Yan Song, and Xiang Wan.
\newblock Generating radiology reports via memory-driven transformer.
\newblock In {\em Proceedings of the Joint Conference of the 59th Annual Meeting of the Association for Computational Linguistics and the 11th International Joint Conference on Natural Language Processing}, August 2021.

\bibitem{wang2022cross}
Jun Wang, Abhir Bhalerao, and Yulan He.
\newblock Cross-modal prototype driven network for radiology report generation.
\newblock In {\em Computer Vision--ECCV 2022: 17th European Conference, Tel Aviv, Israel, October 23--27, 2022, Proceedings, Part XXXV}, pages 563--579. Springer, 2022.

\bibitem{yang2023radiology}
Shuxin Yang, Xian Wu, Shen Ge, Zhuozhao Zheng, S~Kevin Zhou, and Li~Xiao.
\newblock Radiology report generation with a learned knowledge base and multi-modal alignment.
\newblock {\em Medical Image Analysis}, 86:102798, 2023.

\bibitem{irvin2019chexpert}
Jeremy Irvin, Pranav Rajpurkar, Michael Ko, Yifan Yu, Silviana Ciurea-Ilcus, Chris Chute, Henrik Marklund, Behzad Haghgoo, Robyn Ball, Katie Shpanskaya, et~al.
\newblock Chexpert: A large chest radiograph dataset with uncertainty labels and expert comparison.
\newblock In {\em Proceedings of the AAAI conference on artificial intelligence}, volume~33, pages 590--597, 2019.

\bibitem{wang2017chestx}
Xiaosong Wang, Yifan Peng, Le~Lu, Zhiyong Lu, Mohammadhadi Bagheri, and Ronald~M Summers.
\newblock Chestx-ray8: Hospital-scale chest x-ray database and benchmarks on weakly-supervised classification and localization of common thorax diseases.
\newblock In {\em Proceedings of the IEEE conference on computer vision and pattern recognition}, pages 2097--2106, 2017.

\bibitem{shih2019augmenting}
George Shih, Carol~C Wu, Safwan~S Halabi, Marc~D Kohli, Luciano~M Prevedello, Tessa~S Cook, Arjun Sharma, Judith~K Amorosa, Veronica Arteaga, Maya Galperin-Aizenberg, et~al.
\newblock Augmenting the national institutes of health chest radiograph dataset with expert annotations of possible pneumonia.
\newblock {\em Radiology: Artificial Intelligence}, 1(1):e180041, 2019.

\bibitem{yang2025mocoll}
Pu~Yang and Bin Dong.
\newblock Mocoll: Agent-based specific and general model collaboration for image captioning.
\newblock {\em arXiv preprint arXiv:2501.01834}, 2025.

\bibitem{chen2024huatuogpto1medicalcomplexreasoning}
Junying Chen, Zhenyang Cai, Ke~Ji, Xidong Wang, Wanlong Liu, Rongsheng Wang, Jianye Hou, and Benyou Wang.
\newblock Huatuogpt-o1, towards medical complex reasoning with llms, 2024.

\bibitem{radford2021learning}
Alec Radford, Jong~Wook Kim, Chris Hallacy, Aditya Ramesh, Gabriel Goh, Sandhini Agarwal, Girish Sastry, Amanda Askell, Pamela Mishkin, Jack Clark, et~al.
\newblock Learning transferable visual models from natural language supervision.
\newblock In {\em International conference on machine learning}, pages 8748--8763. PmLR, 2021.

\bibitem{christiano2017deep}
Paul~F Christiano, Jan Leike, Tom~B Brown, Miljan Martic, Shane Legg, and Dario Amodei.
\newblock Deep reinforcement learning from human preferences.
\newblock {\em arXiv preprint arXiv:1706.03741}, 2017.

\bibitem{schulman2017proximal}
John Schulman, Filip Wolski, Prafulla Dhariwal, Alec Radford, and Oleg Klimov.
\newblock Proximal policy optimization algorithms.
\newblock {\em arXiv preprint arXiv:1707.06347}, 2017.

\bibitem{shao2024deepseekmathpushinglimitsmathematical}
Zhihong Shao, Peiyi Wang, Qihao Zhu, Runxin Xu, Junxiao Song, Xiao Bi, Haowei Zhang, Mingchuan Zhang, Y.~K. Li, Y.~Wu, and Daya Guo.
\newblock Deepseekmath: Pushing the limits of mathematical reasoning in open language models, 2024.

\bibitem{lai2025med}
Yuxiang Lai, Jike Zhong, Ming Li, Shitian Zhao, and Xiaofeng Yang.
\newblock Med-r1: Reinforcement learning for generalizable medical reasoning in vision-language models.
\newblock {\em arXiv preprint arXiv:2503.13939}, 2025.

\bibitem{liu2025visual}
Ziyu Liu, Zeyi Sun, Yuhang Zang, Xiaoyi Dong, Yuhang Cao, Haodong Duan, Dahua Lin, and Jiaqi Wang.
\newblock Visual-rft: Visual reinforcement fine-tuning.
\newblock {\em arXiv preprint arXiv:2503.01785}, 2025.

\bibitem{huang2025vision}
Wenxuan Huang, Bohan Jia, Zijie Zhai, Shaosheng Cao, Zheyu Ye, Fei Zhao, Zhe Xu, Yao Hu, and Shaohui Lin.
\newblock Vision-r1: Incentivizing reasoning capability in multimodal large language models.
\newblock {\em arXiv preprint arXiv:2503.06749}, 2025.

\bibitem{loshchilov2019decoupledweightdecayregularization}
Ilya Loshchilov and Frank Hutter.
\newblock Decoupled weight decay regularization, 2019.

\bibitem{10.1145/3394486.3406703}
Jeff Rasley, Samyam Rajbhandari, Olatunji Ruwase, and Yuxiong He.
\newblock Deepspeed: System optimizations enable training deep learning models with over 100 billion parameters.
\newblock In {\em Proceedings of the 26th ACM SIGKDD International Conference on Knowledge Discovery \& Data Mining}, KDD '20, page 3505–3506, New York, NY, USA, 2020. Association for Computing Machinery.

\bibitem{zhao2023pytorchfsdpexperiencesscaling}
Yanli Zhao, Andrew Gu, Rohan Varma, Liang Luo, Chien-Chin Huang, Min Xu, Less Wright, Hamid Shojanazeri, Myle Ott, Sam Shleifer, Alban Desmaison, Can Balioglu, Pritam Damania, Bernard Nguyen, Geeta Chauhan, Yuchen Hao, Ajit Mathews, and Shen Li.
\newblock Pytorch fsdp: Experiences on scaling fully sharded data parallel, 2023.

\bibitem{zheng2025easyr1}
Yaowei Zheng, Junting Lu, Shenzhi Wang, Zhangchi Feng, Dongdong Kuang, and Yuwen Xiong.
\newblock Easyr1: An efficient, scalable, multi-modality rl training framework.
\newblock \url{https://github.com/hiyouga/EasyR1}, 2025.

\bibitem{10.1145/3689031.3696075}
Guangming Sheng, Chi Zhang, Zilingfeng Ye, Xibin Wu, Wang Zhang, Ru~Zhang, Yanghua Peng, Haibin Lin, and Chuan Wu.
\newblock Hybridflow: A flexible and efficient rlhf framework.
\newblock In {\em Proceedings of the Twentieth European Conference on Computer Systems}, EuroSys '25, page 1279–1297, New York, NY, USA, 2025. Association for Computing Machinery.

\end{thebibliography}

\clearpage

\section{Supplementary}
\setcounter{figure}{0} % 在该节开头重置
\renewcommand{\figurename}{Supplementary Figure} % 重定义名称
\setcounter{table}{0} % 在该节开头重置
\renewcommand{\tablename}{Supplementary Table} % 重定义名称

\setcounter{algorithm}{0} % 重置算法计数器
\subsection{Training Details}

We employ Qwen2-VL-7B as our baseline model, chosen for its strong performance in vision-language tasks and efficient parameter utilization. As detailed in Supplementary Table~\ref{tab:hyperparams-combined}(a), for Stage 1 of DiagCoT, we utilize the AdamW~\cite{loshchilov2019decoupledweightdecayregularization} optimizer with beta1 and beta2 coefficients set to 0.9 and 0.99, respectively. Additionally, we adopt a cosine learning rate schedule with an initial learning rate of 2e-6, a batch size of 64 samples, a weight decay of 0.0, and a gradient clipping threshold of 1. The experiments are conducted on 2 A800 GPUs using DeepSpeed-ZeRO3~\cite{10.1145/3394486.3406703}. In this stage, the model is trained for 3485 steps, taking approximately 2 days to complete.As shown in Supplementary Table~\ref{tab:hyperparams-combined}(b), Stage 2 differs from Stage 1 primarily in batch size and learning rate, which are set to 16 and 1e-5, respectively. In this stage, the model is trained for 1136 steps, taking approximately 9 hours to complete. For the third stage, which employs reinforcement learning enhanced with formate and accuracy rewards, the detailed training configuration is summarized in Supplementary Table~\ref{tab:hyperparams-stage3}. The optimizer configuration remains consistent with the previous stage, utilizing a learning rate of 1e-6. Both the global batch size and the mini-batch update size are configured at 1. The KL divergence loss coefficient is set to 0.05, with a group number (G) of 8. The reward function integrates incentives for output structure, result accuracy. All experiments are executed on a cluster of 2 A800 GPUs, implemented via PyTorch Fully Sharded Data Parallel (FSDP)~\cite{zhao2023pytorchfsdpexperiencesscaling} and the  EasyR1~\cite{zheng2025easyr1} based on the VeRL~\cite{10.1145/3689031.3696075} framework. This stage involves training the model for 435 steps, requiring approximately 4 days to finish.

\begin{table}[htbp]
  \centering
  \caption{Training hyper-parameters in training stage 1(left) and Stage 2(right)}
  \begin{subtable}[b]{0.48\textwidth}
    \centering
    \begin{tabular}{l>{\raggedleft\arraybackslash}p{2cm}}
      \toprule
      Parameter name & Value \\
      \midrule
      Global batchsize & 64 \\
      Learning rate & 3e-4 \\
      Optimizer & AdamW \\
      $\beta_{1}$ of optimizer & 0.9 \\
      $\beta_{2}$ of optimizer & 0.99 \\
      Warmup steps & 280 \\
      Learning rate scheduler & Cosine \\
      Gradient clipping threshold & 1.0 \\
      Weight decay & 0.0 \\
      Platform & 2*A800 \\
      Training steps & 3,485 steps \\
      Training times & about 2 days \\
      DeepSpeed & zero3 \\
      \bottomrule
    \end{tabular}
    \caption{}
    \label{tab:hyperparams-left}
  \end{subtable}
  \hfill
  \begin{subtable}[b]{0.48\textwidth}
    \centering
    \begin{tabular}{l>{\raggedleft\arraybackslash}p{2cm}}
      \toprule
      Parameter name & Value \\
      \midrule
      Global batchsize & 16 \\
      Learning rate & 1e-5 \\
      Optimizer & AdamW \\
      $\beta_{1}$ of optimizer & 0.9 \\
      $\beta_{2}$ of optimizer & 0.99 \\
      Warmup step & 0.0 \\
      Learning rate scheduler & Cosine \\
      Gradient clipping threshold & 1.0 \\
      Weight decay & 0.0 \\
      Platform & 2*A800 \\
      Training steps & 1,136 steps \\
      Training times & about 9 hrs \\
      DeepSpeed & zero3 \\
      \bottomrule
    \end{tabular}
    \caption{}
    \label{tab:hyperparams-right}
  \end{subtable}
  \label{tab:hyperparams-combined}
\end{table}

\begin{table}[htbp]
\centering
\caption{Training hyper-parameters in training stage 3.}
\begin{tabular}{lc}
\toprule
Parameter name & Value \\
\midrule
Global batchsize & 1 \\
Updating batchsize & 1 \\
KL loss coefficient & 0.05 \\
Gradient clipping threshold & 1.0 \\
% Entropy loss coefficient & xxx \\
Group number (G) & 8 \\
Learning rate & 1e-6 \\
Optimizer & AdamW \\
$\beta_{1}$ of optimizer & 0.9 \\
$\beta_{2}$ of optimizer & 0.99 \\
Platform & 2*A800 \\
Training steps & 435 steps \\
Training times & about 4 days \\
Distributed platform & FSDP without offloading \\
Reward function & outcome format and outcome accuracy \\
\bottomrule
\end{tabular}
\label{tab:hyperparams-stage3}
\end{table}

\subsection{CoT Collection Prompt Design}
\label{sec: CoT Collection Prompt Design}
\begin{figure*}[!htbp]
\begin{center}
\includegraphics[width=\textwidth]{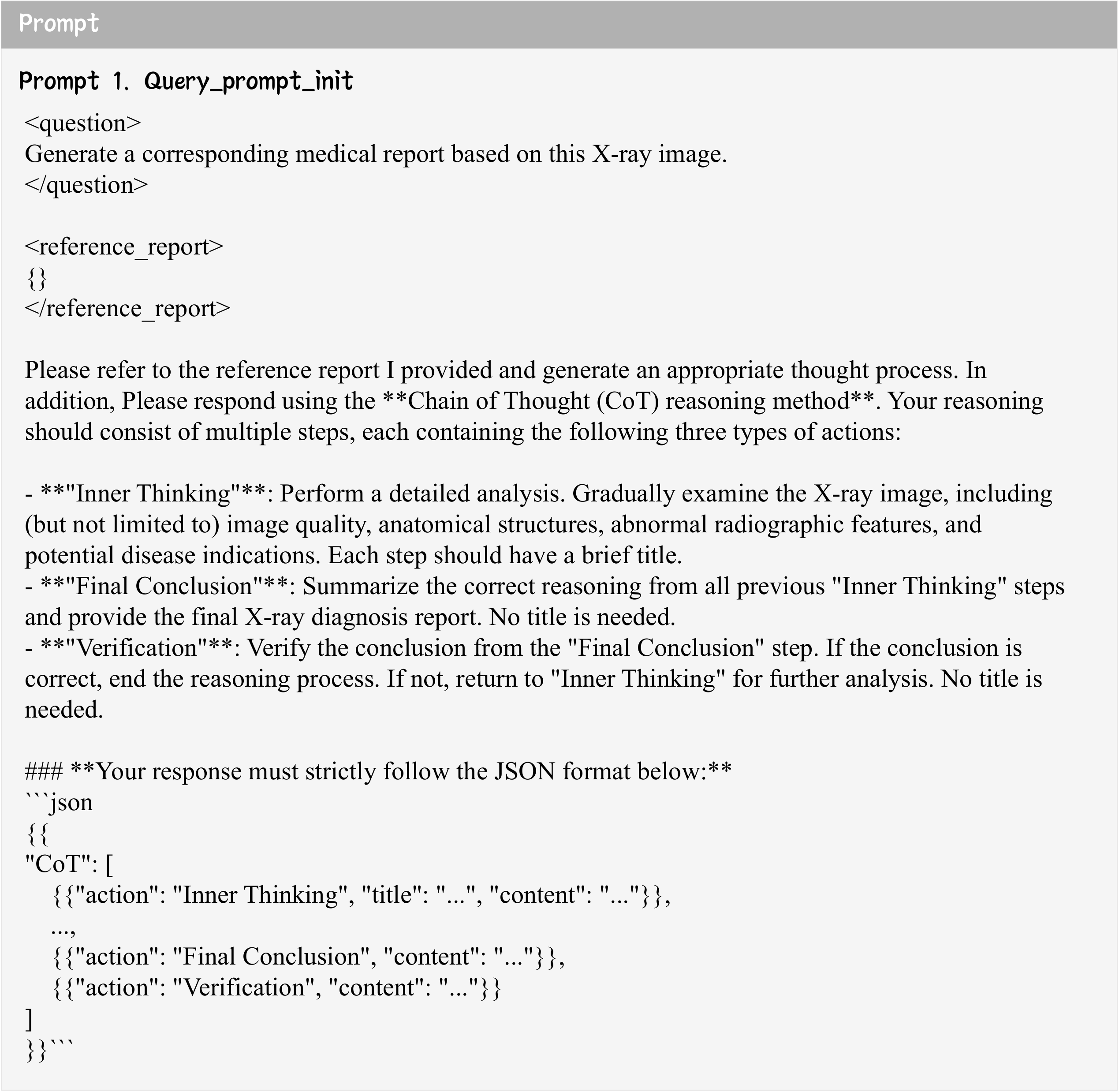}
\end{center}

\caption{\textbf{Ptompt 1. }The initial prompt aims to generate the model's complete and preliminary reasoning process for the medical image without employing any guided strategies, serving as a baseline.}
\label{fig: Supplementary-cot-collection_1}
  % \vspace{-.15in}
\end{figure*}

\begin{figure*}[!htbp]
\begin{center}
\includegraphics[width=\textwidth]{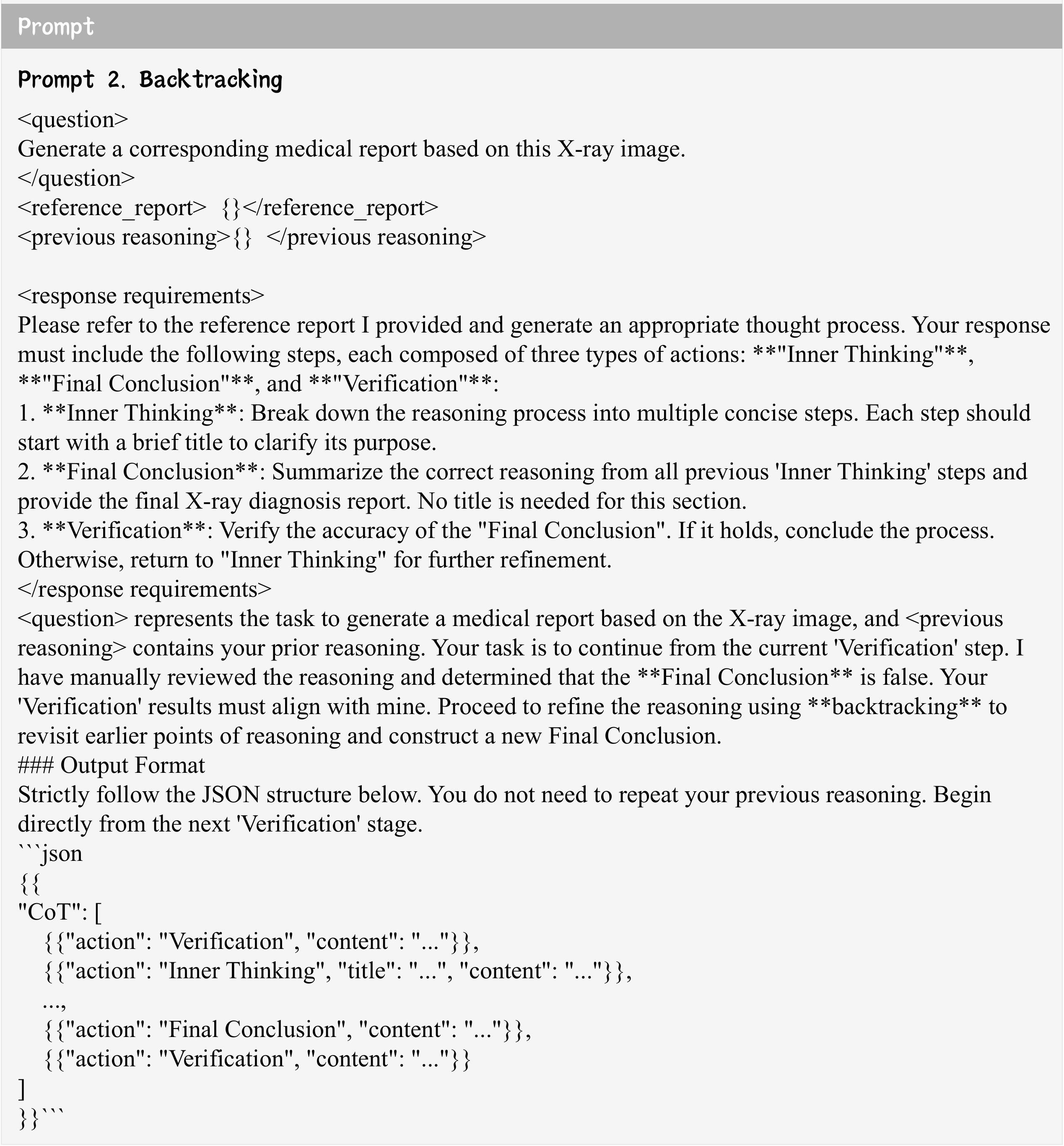}
\end{center}

\caption{\textbf{Ptompt 2. }This figure illustrates the Backtracking Prompt Strategy, which guides the model to trace back through its reasoning path to identify and correct potential errors in earlier steps.}
\label{fig: Supplementary-cot-collection_2}
  % \vspace{-.15in}
\end{figure*}

\begin{figure*}[!htbp]
\begin{center}
\includegraphics[width=\textwidth]{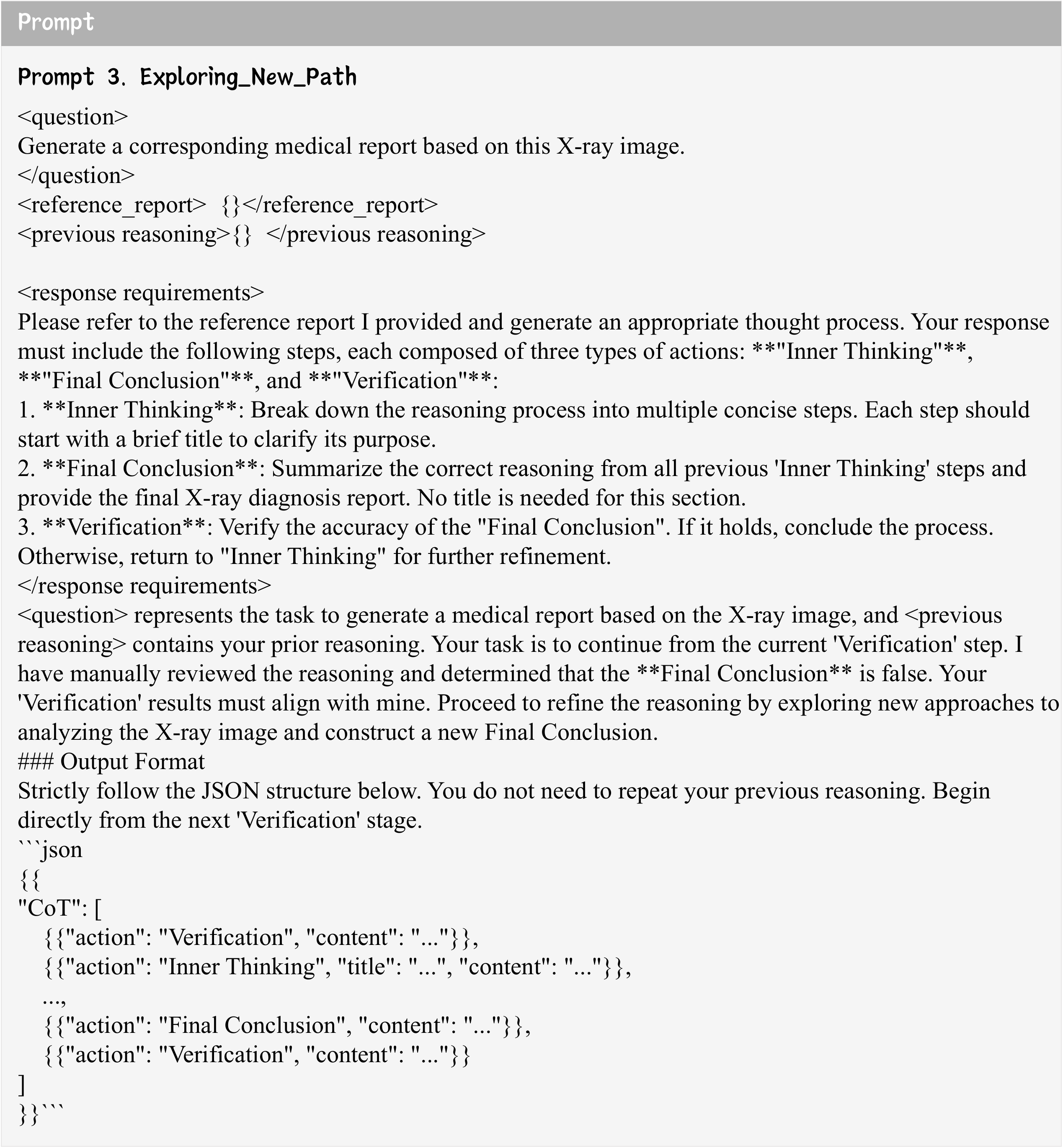}
\end{center}

\caption{\textbf{Ptompt 3. }This figure outlines the Exploration Prompt Strategy, designed to steer the model towards considering alternative diagnostic hypotheses or reasoning paths to broaden its analysis.}
\label{fig: Supplementary-cot-collection_3}
  % \vspace{-.15in}
\end{figure*}

\begin{figure*}[!htbp]
\begin{center}
\includegraphics[width=\textwidth]{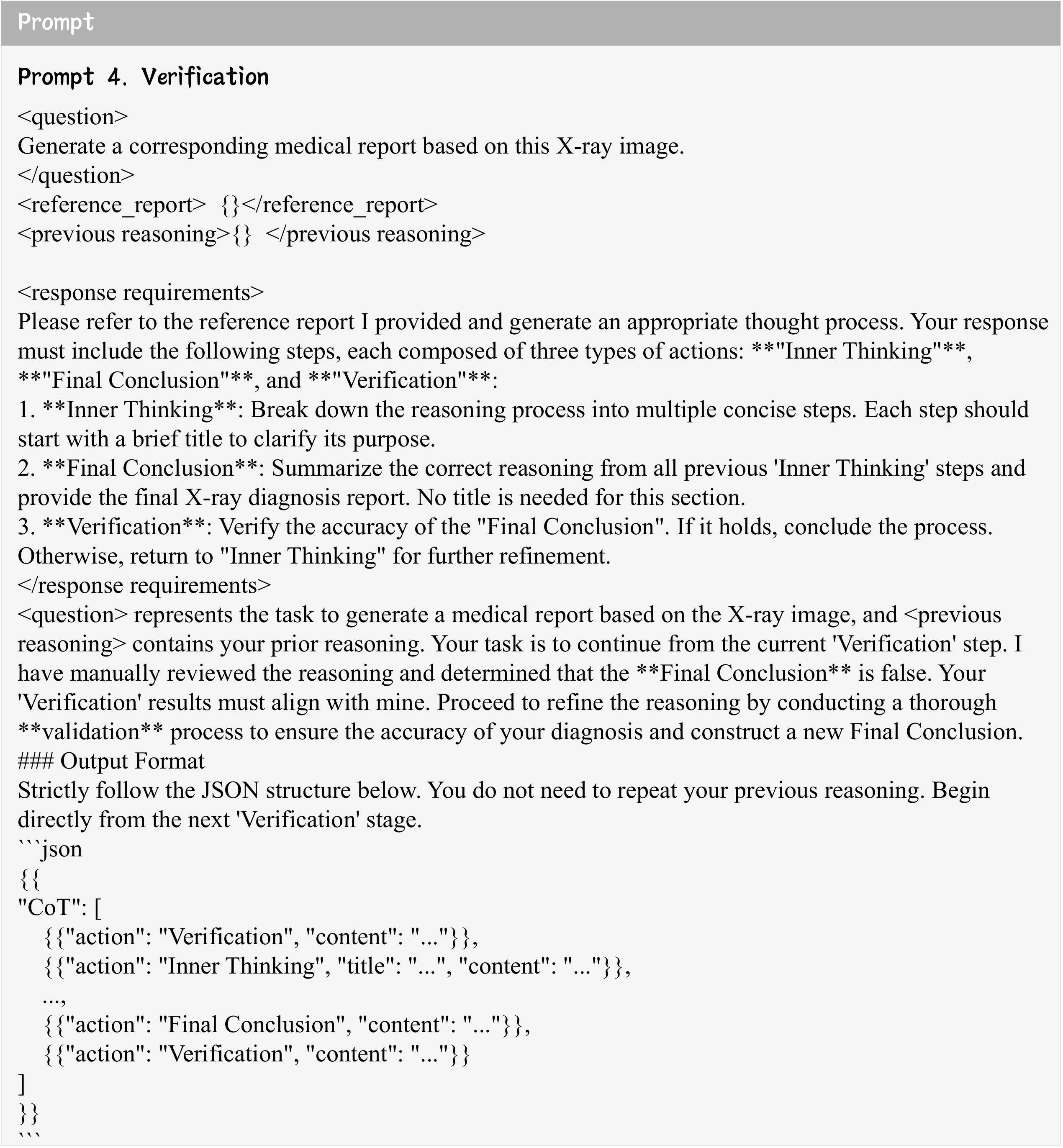}
\end{center}

\caption{\textbf{Ptompt 4. }This figure demonstrates the Verification Prompt Strategy, which instructs the model to self-assess and confirm the confidence level and evidentiary support for its preliminary conclusions.}
\label{fig: Supplementary-cot-collection_4}
  % \vspace{-.15in}
\end{figure*}

\begin{figure*}[!htbp]
\begin{center}
\includegraphics[width=\textwidth]{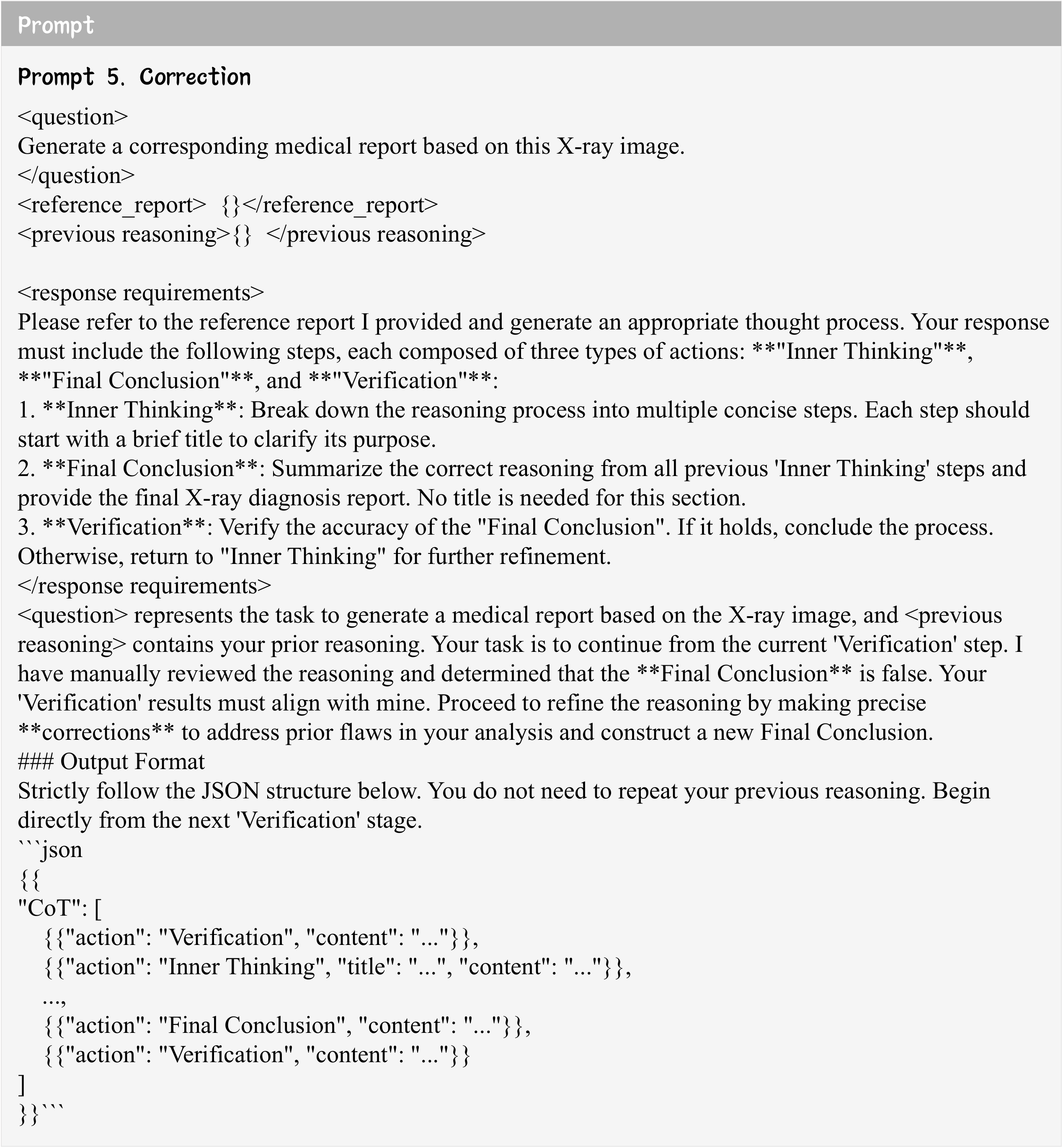}
\end{center}

\caption{\textbf{Ptompt 5. }This figure describes the Correction Prompt Strategy, which directly instructs the model to identify specific errors within its current reasoning and provide a corrected output.}
\label{fig: Supplementary-cot-collection_5}
  % \vspace{-.15in}
\end{figure*}

\begin{figure*}[!htbp]
\begin{center}
\includegraphics[width=\textwidth]{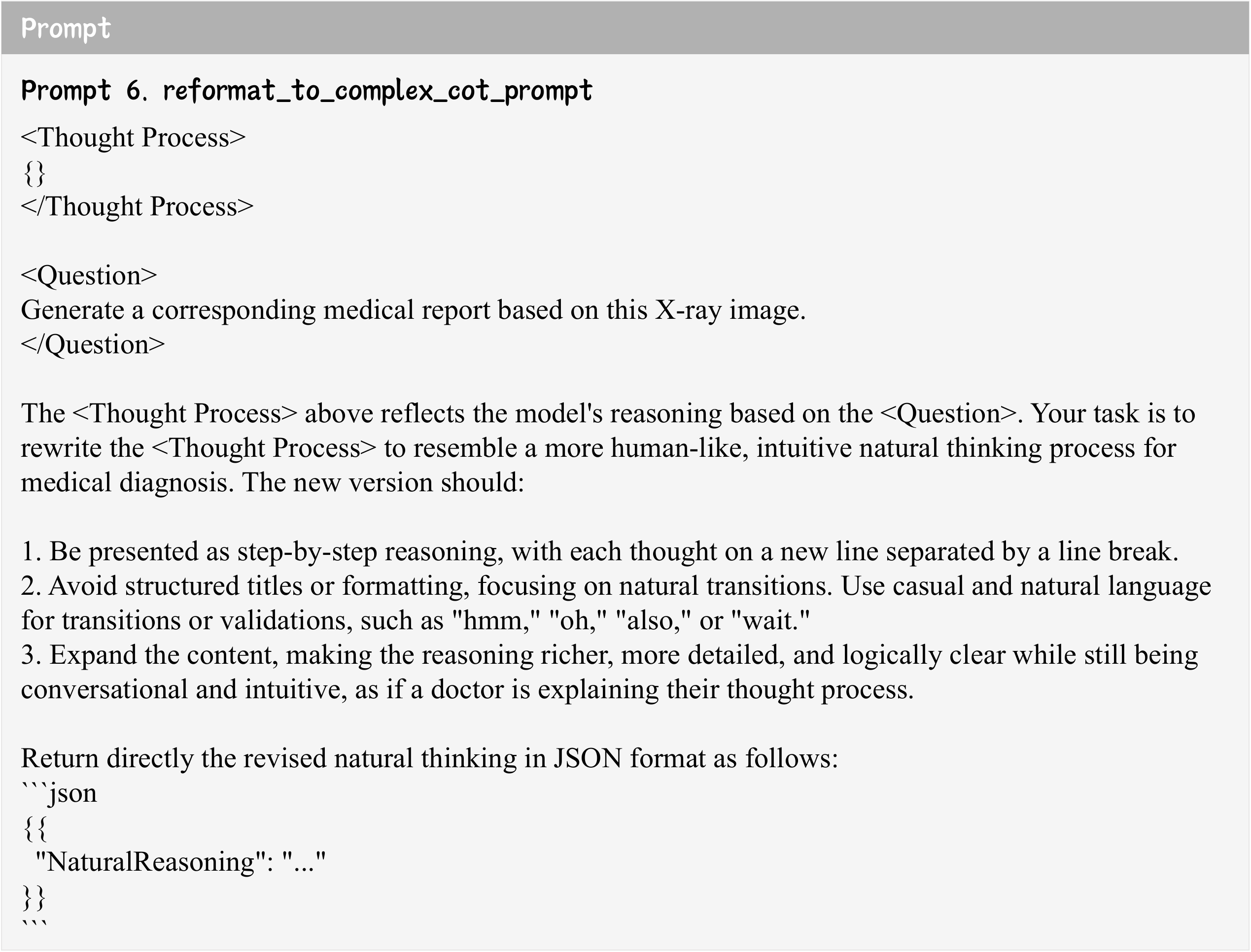}
\end{center}

\caption{\textbf{Ptompt 6. }This figure presents the Natural Language Conversion Prompt, whose function is to translate the model's internal structured chain of thought into a fluent and easily understandable natural language paragraph.
}
\label{fig: Supplementary-cot-collection_6}
  % \vspace{-.15in}
\end{figure*}

This section details the prompt designs developed for Chain-of-Thought (CoT) collection during the CoT tuning stage, as visually summarized across Supplementary Figures~\ref{fig: Supplementary-cot-collection_1} to~\ref{fig: Supplementary-cot-collection_6}. The collection protocol comprises five distinct prompt strategies, each designed to elicit a specific reasoning behavior from the model:
\begin{itemize}
\item The initial prompt (Supplementary Figure~\ref{fig: Supplementary-cot-collection_1}) aims to generate the model's complete and preliminary reasoning process for the medical image without employing any guided strategies, serving as a baseline.
\item The four subsequent strategies (Supplementary Figures~\ref{fig: Supplementary-cot-collection_2} to~\ref{fig: Supplementary-cot-collection_5}) are designed to guide the model—through backtracking, exploring new paths, verification, and correction—toward producing more reliable and robust CoT data.
\item A fifth and final prompt (Supplementary Figure~\ref{fig: Supplementary-cot-collection_6}) is dedicated to translating complex chains of thought into natural language easily understandable by humans, thereby significantly enhancing the intuitiveness and readability of the finalized CoT data.
\end{itemize}

\subsection{Complete Prediction Results Visualization}
\label{sec: Complete Prediction Results Visualization}
\begin{figure*}[!htbp]
\begin{center}
\includegraphics[width=\textwidth]{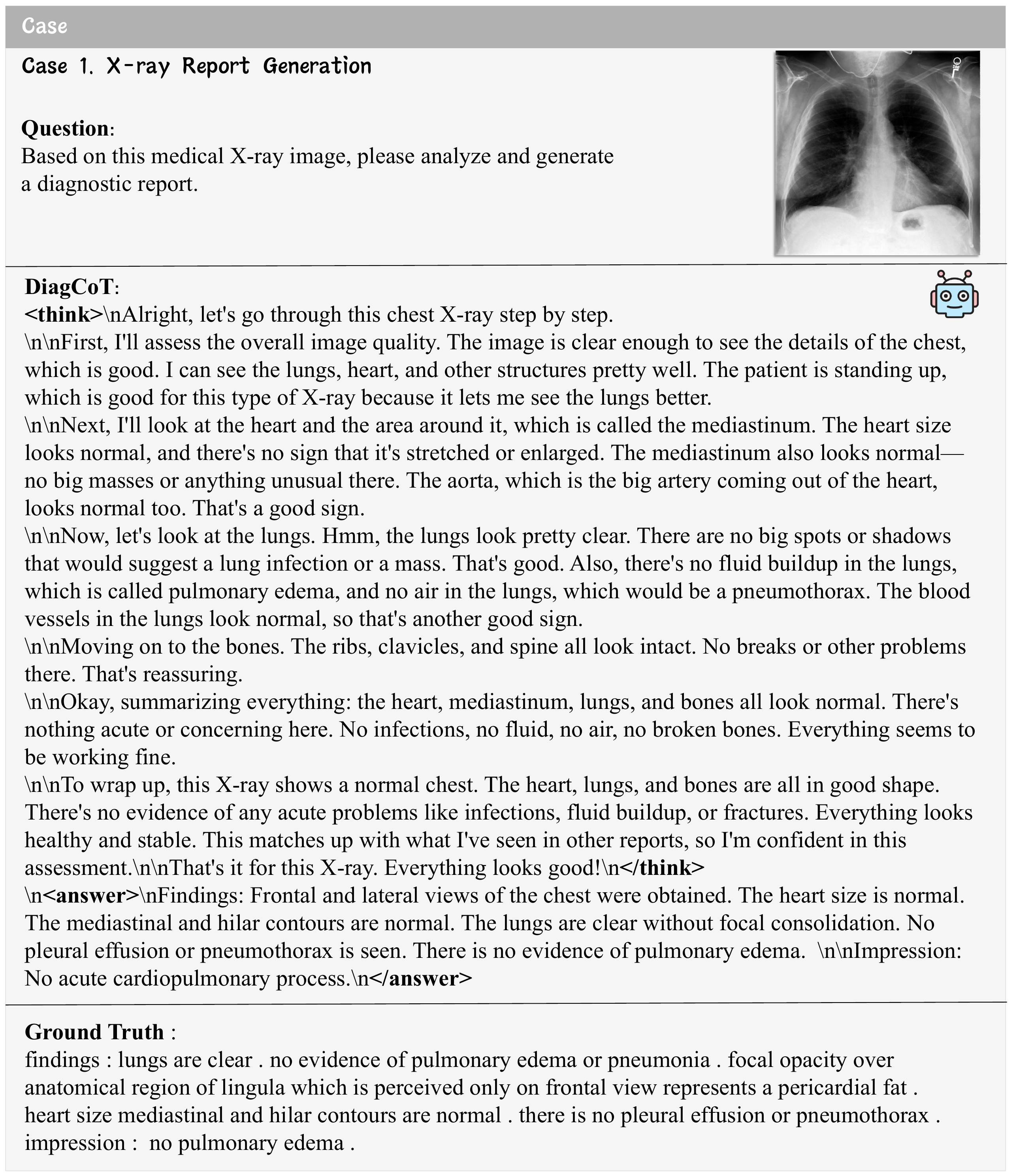}
\end{center}

\caption{\textbf{Case 1. }It illustrates the reasoning process of DiagCoT in the X-ray Report Generation task, demonstrating how the model conducts a detailed, multi-perspective analysis of the image to derive the final diagnostic report.
}
\label{fig: Supplementary-model-pre-cot_1}
  % \vspace{-.15in}
\end{figure*}

\begin{figure*}[!htbp]
\begin{center}
\includegraphics[width=\textwidth]{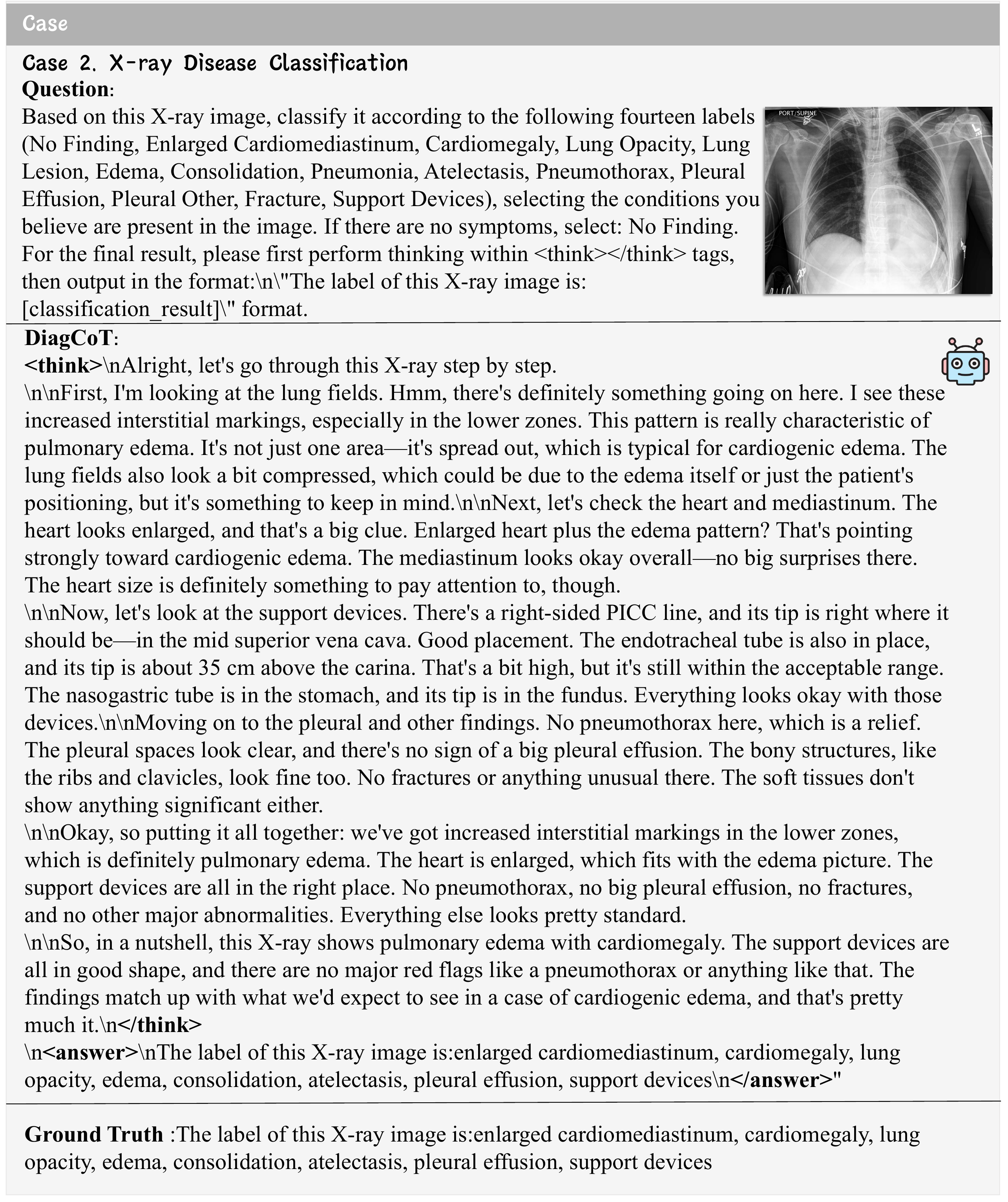}
\end{center}

\caption{\textbf{Case 2. }It showcases DiagCoT’s performance on the Disease Classification task. The model is provided with a predefined set of labels and performs the task in a multiple-choice manner. It engages in deliberate reasoning before selecting the final classification result.
}
\label{fig: Supplementary-model-pre-cot_2}
  % \vspace{-.15in}
\end{figure*}

\begin{figure*}[!htbp]
\begin{center}
\includegraphics[width=\textwidth]{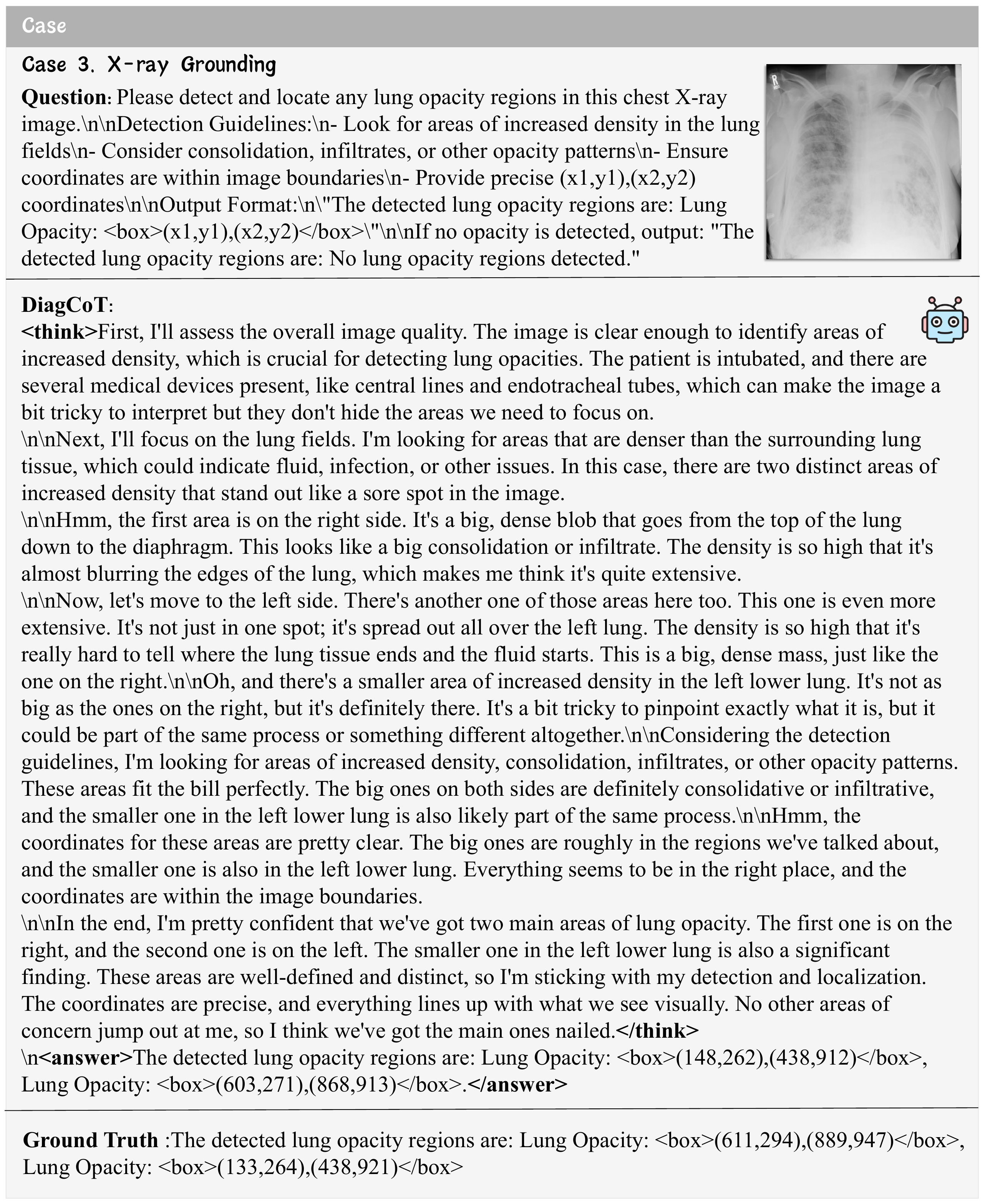}
\end{center}

\caption{\textbf{Case 3. }It presents DiagCoT’s performance on the Pathology Grounding task. the model performs comprehensive analysis to accurately predict the top-left and bottom-right coordinates of the region of interest.
}
\label{fig: Supplementary-model-pre-cot_3}
  % \vspace{-.15in}
\end{figure*}

This section provides a detailed visualization of DiagCoT's reasoning process across three tasks: X-ray Report Generation, Disease Classification, and Pathology Grounding. 
\begin{itemize}
    \item Supplementary Figure~\ref{fig: Supplementary-model-pre-cot_1} illustrates DiagCoT's performance in X-ray Report Generation. It can be observed that the model first conducts a detailed, multi-perspective analysis of the medical image before deriving the final diagnostic report.
    \item Supplementary Figure~\ref{fig: Supplementary-model-pre-cot_2} demonstrates DiagCoT's approach to Disease Classification. The model is first prompted with the available label set for the dataset, framing the task as a multiple-choice selection. As shown, it also undergoes thorough reasoning before determining the final classification result.
    \item Supplementary Figure~\ref{fig: Supplementary-model-pre-cot_3} presents DiagCoT's performance in Pathology Grounding. Similarly, the model performs comprehensive analysis to arrive at the final coordinates for the top-left and bottom-right bounding boxes.
\end{itemize}

\subsection{Complete Experimental Results Table}
\label{sec: Complete Experimental Results Table}
Evaluation of X-ray report generation:
\begin{itemize}
    \item X-Ray report generation within-dataset evaluation on the MIMIC-CXR dataset: Supplementary Table~\ref{table1:rrg_within}.
    \item X-Ray report generation cross-dataset evaluation on the IU-Xray dataset. Supplementary Table~\ref{table2:rrg_cross}.
\end{itemize}

Evaluation of Diease classification: Supplementary Table~\ref{tab:Diease cls}.

Evaluation of Pathology grounding: Supplementary Table~\ref{tab:Pathology grounding}.

Ablation on Training Strategy: Supplementary Table~\ref{tab:Ablation1}.

\begin{table}[h]
\centering
\caption{X-Ray report generation within-dataset evaluation on the MIMIC-CXR dataset.  \textbf{Bold} represents the optimal performance}
\setlength{\tabcolsep}{5pt} 
\begin{tabularx}{\textwidth}{lccccccc}
\toprule
\text{Model} & \text{BLEU1} & \text{BLEU2} & \text{BLEU3} & \text{BLEU4} & \text{ROUGE-L} & \text{METEOR} & \text{CIDEr} \\
\midrule
\text{R2gen} &  0.2750 & 0.1667 &  0.1132 & 0.0813  & 0.2634 & 0.1288 & 0.1138 \\
\text{R2genCMN} &  0.2780 & 0.1625 &  0.1076 & 0.0757  & 0.2136 & 0.1281 & 0.0197 \\
\text{XproNet} &  0.2686 & 0.1586 &  0.1021 & 0.0639  & 0.2565 & 0.1392 & 0.1247 \\
\text{M2KT} &  0.2377 & 0.1308 &  0.0811 & 0.0544  & 0.2096 & 0.1065 & 0.0565 \\

\midrule
\text{LLAVA-Med} & 0.1069 & 0.0324 & 0.0032   & 0.0005 & 0.0984& 0.0521 & 0.0043 \\
\text{CXR-LLAVA} & 0.1661 & 0.0863 & 0.0403 & 0.0175 & 0.1716 & 0.0922 & 0.0257 \\
\text{ChestX-Reasoner} & 0.0612 & 0.0242 & 0.0074 & 0.0018 & 0.0770 & 0.0561 & 0.0009 \\
\text{CheXagent-2-3b} & 0.0147 & 0.0082 & 0.0042 & 0.0017 & 0.0829 & 0.0322 & 0.0013 \\
\text{Deepseek-VL-7B-chat} & 0.1013 & 0.0343 & 0.0076 & 0.0014 & 0.1174 & 0.0055 & 0.0624\\
\text{Qwen2.5-VL-32B-Instruct} & 0.0797 & 0.0359 & 0.0143 & 0.0024 & 0.0895 & -- & 0.1050 \\
\text{Qwen2.5-VL-72B-Instruct-AWQ} & 0.1937 & 0.0676 & 0.0151 & 0.0028 & 0.1467 & 0.0153 & 0.0900\\
\midrule
\text{Qwen2-VL-7B-Base} & 0.1090 & 0.0397 & 0.0120 & 0.0030 & 0.1053 & 0.0950 & --     \\
\text{Qwen2.5-VL-32B-LoRA} & 0.2290 & 0.1310& 0.0771 & 0.0458 & 0.2351& 0.0744 & \textbf{0.1285}\\
\text{DiagCoT(Stage 1)} & 0.1888 & 0.1146 & 0.0743 & 0.0471 & 0.2098 & 0.1077 & 0.0647 \\
\text{DiagCoT(Stage 2)}& 0.2524 & 0.1326 & 0.0715 & 0.0388 & 0.2220 & 0.1365 & 0.0663 \\
% \text{Qwen2-VL-7B-RFT(Stage 3)} & 0.3416 & 0.2075 & 0.1342 & 0.0854 & 0.2703 & 0.1698 & 0.0973 \\
\text{DiagCoT(Stage 3)} & \textbf{0.3260} & \textbf{0.2024} & \textbf{0.1343} & \textbf{0.0900} & \textbf{0.2717}& \textbf{0.1619} & 0.1152 \\

\bottomrule
\label{table1:rrg_within}
\end{tabularx}
\end{table}

\begin{table}[h]
\centering
\caption{X-Ray report generation cross-dataset evaluation on the IU-Xray dataset.  \textbf{Bold} represents the optimal performance.}
\setlength{\tabcolsep}{5pt} 
\begin{tabularx}{\textwidth}{lccccccc}
\toprule
\text{Model} & \text{BLEU1} & \text{BLEU2} & \text{BLEU3} & \text{BLEU4} & \text{ROUGE-L} & \text{METEOR} & \text{CIDEr} \\
\midrule
\text{R2gen} &  0.2928 & 0.1587 &  0.0949 & 0.0570  & 0.2462 &  0.1602 & 0.0762 \\
\text{R2genCMN} &  0.1364 & 0.0749 &  0.0453 & 0.0278  & 0.1593 &  0.1299 & 0.0002 \\
\text{M2KT} & 0.2655 & 0.1385 &  0.0789 & 0.0441  & 0.2221 & 0.1516 & 0.0243 \\
\text{XproNet} &  0.3023 & 0.1655 &  0.1001 & 0.0607  & 0.2588 & 0.1693 &  0.0779 \\
\midrule
\text{LLAVA-Med} & 0.1018 & 0.0220 & 0.0028 & 0.0003  & 0.0995 & 0.0575 & 0.0118 \\
\text{CXR-LLAVA} & 0.2637 & 0.1206 & 0.0515 & 0.0170 & 0.2163 & 0.1434 & 0.0685 \\
\text{ChestX-Reasoner} & 0.0696 & 0.0301 & 0.0140 & 0.0077 & 0.1059 & 0.1453 & 0.0037 \\
\text{CheXagent-2-3b} & 0.0408 & 0.0178 & 0.0106 & 0.0068 & 0.1014 & 0.0650 & 0.0046 \\
\text{Deepseek-VL-7B-chat} & 0.1248 & 0.0299 & 0.0054 & 0.0007 & 0.1098 & 0.0207 & 0.0694 \\
\text{Qwen2.5-VL-32B-Instruct} & 0.1181 & 0.0466 & 0.0164 & 0.0039 & 0.1283 & 0.0005 & 0.1139 \\
\text{Qwen2.5-VL-72B-Instruct-AWQ} & 0.1750 & 0.0639 & 0.0221 & 0.0044 & 0.1644 & 0.0113 & 0.1170 \\
\midrule
\text{Qwen2-VL-7B-Base} & 0.0292 & 0.0111 & 0.0031 & 0.0007 & 0.0280 & 0.0206 & 0.0015     \\
\text{Qwen2.5-VL-32B-LoRA} & 0.1558 & 0.0691 & 0.0287 & 0.0100 & 0.1593 & 0.0046 & \textbf{0.1384}\\
\text{DiagCoT(Stage 1)} & 0.3020 & 0.1887 & \textbf{0.1221} & \textbf{0.0626} & 0.2652 & 0.1569 & 0.1283 \\
\text{DiagCoT(Stage 2)}& 0.3101 & 0.1857 & 0.1108 & 0.0569 & 0.2645 &  0.1766 & 0.1041 \\
% \text{Qwen2-VL-7B-RFT(Stage 3)} & 0.3435 & 0.2234 & 0.1346 & 0.0750 & 0.3182 & 0.1876 & 0.3140\\
\text{DiagCoT(Stage 3)} & \textbf{0.3363} & 
\textbf{0.1930} & 0.1096 & 0.0566 & \textbf{0.3005} & \textbf{0.2110} & 0.0681\\

\bottomrule
\label{table2:rrg_cross}
\end{tabularx}
\end{table}

\begin{table}
\caption{Diease classification performance (AUC: \%) on the CheXpert dataset. \textbf{Bold} represents the optimal performance, while \underline{Text} exhibits suboptimal performance.}
% \resizebox{\textwidth}{!}{ % 等比例缩放到文本宽度
\setlength{\tabcolsep}{1.7pt}
\begin{tabular}{l*{15}{c}} % 16列（第一列左对齐+15列居中）
\toprule
Model & Enl. & Car. & Opa. & Ede. & Con. & Ate. & Ple. & Sup. & Pne. & Les. & Pne2. & Ple2. & Fra. & Nofi. & Avg.\\
\midrule
 Qwen2-VL-7B-Base & 52.1 & 55.3 & 44.0 & 59.8 & 50.5 & 54.8 & 52.4 & 61.5 & 50.4 & 51.1 & 48.7 & 48.0 & 49.3 & 53.3  & 52.2 \\
 CXR-LLaVA & 51.7 & 54.0 & 50.3 & 52.8 & 47.2 & 57.7 & 46.2 & 50.0 & 46.9 & 50.0 & 52.6 & 50.0 & 50.1 & 56.2 & 51.1  \\
 ChestX-Reasoner-7B & 50.0 & 52.9  & 66.2 & \textbf{69.7} & 50.6 & 51.5 & 67.1 & 66.2 & 50.4 & 50.4 & 51.5 & 49.8 & 50.2 & 76.3 &  57.3 \\
 CheXagent-2-3b & 50.8 &64.1  & \textbf{79.4} & 65.8 & 52.2 & 58.5 & \textbf{79.3} & \textbf{69.0} & 53.5 & \textbf{55.6} & \textbf{69.9} & 49.3 & 49.3 & \textbf{81.2} &  62.7 \\
 Deepseek-VL-7b-chat & 50.0 & 51.8 & 50.8 & 50.0 & 50.0 & 50.1 & 50.0 & 50.1 & 50.0 & 49.8 & 50.0 & 50.0 & 50.0 & 51.4 & 50.3 \\
 Qwen2.5-VL-32B-Instruct & 53.3 & 52.6 & 53.5 & 51.1 & 50.8 & 50.3 & 52.9 & 66.9 & 51.9 & 50.1 & 49.8 & 51.2 & 50.3 & 59.3 &  53.1 \\
 Qwen2.5-VL-72B-AWQ & 58.7 & 58.6 & 57.3 & 54.8 & 53.1 & 50.6 & 51.7& 67.6 & \textbf{54.3} & 50.6 &  50.0 & 50.0 & 50.0 & 58.0 & 54.7 \\
 DiagCoT & \textbf{68.5} & \textbf{65.3} & \underline{71.0} & \underline{65.9} & \textbf{65.0} & \textbf{65.1} & 62.4 & 63.6 & 50.8 & 50.0 & 50.0 & \textbf{66.4} & \textbf{58.1} & \underline{77.3} & \textbf{62.8} \\
\bottomrule
\end{tabular}
\label{tab:Diease cls}
\end{table}

%Groungding原先的表格
\begin{table}
\centering
\caption{Pathology grounding evaluation on the RSNA dataset.  \textbf{Bold} represents the optimal performance.} 
\setlength{\tabcolsep}{23pt}
\begin{tabular}{l*{2}{c}} % 7列全部居中
\toprule
Model &  ACC & mIoU \\ % 表头行
\midrule
Qwen2-VL-7B-Base & 0.0879  & 0.1356 \\
 ChestX-Reasoner-7B &  0.3463 & 0.0757 \\ % 数据行1
 Deepseek-VL-7b-chat  & 0.2161 &  0.0913 \\ % 数据行2
 Qwen2.5-VL-32B-Instruct  & 0.2959 & 0.1455 \\ % 数据行3
 Qwen2.5-VL-72B-AWQ  & 0.3604 & 0.1497 \\ % 数据行4
 DiagCoT & \textbf{0.7307} & \textbf{0.3073} \\ % 数据行5 目前采用的是基于报告的Stage2做的，merger + LLM 训的
\bottomrule
\end{tabular}
\label{tab:Pathology grounding} % 可选标签
\end{table}

\begin{table}[h]
\centering
\caption{Ablation on Training Strategy. \textbf{Bold} represents the optimal performance.}
\setlength{\tabcolsep}{8pt} 
\begin{tabularx}{\textwidth}{lccccccc}
\toprule
\text{Model} & \text{BLEU1} & \text{BLEU2} & \text{BLEU3} & \text{BLEU4} & \text{ROUGE-L} & \text{METEOR} & \text{CIDEr} \\
\midrule
\text{DiagCoT-SFT(CoT)} & 0.2096 & 
0.1130 & 0.0651 & 0.0390 & 0.2184 & 0.0653 & 0.1189\\
\text{DiagCoT-RL(O)} & 0.2652 & 
0.1110 & 0.0488 & 0.0229 & 0.1927 & 0.0262 & 0.1160\\
\text{DiagCoT-RL(W.1)} & 0.2818 & 
0.1697 & 0.1071 & 0.0652 & 0.2395 & 0.0559 & 0.1433 \\
\text{DiagCoT-RL(W.2)} & 0.3128 & 
0.1900 & 0.1243 & 0.0853 &  0.2661 & 0.1118 & \textbf{0.1553}\\

\text{DiagCoT} & \textbf{0.3260} & 
\textbf{0.2024} & \textbf{0.1343} & \textbf{0.0900} & \textbf{0.2717} & \textbf{0.1619} & 0.1152\\
\bottomrule
\label{tab:Ablation1}
\end{tabularx}
\end{table}

\subsection{Algorithm}

This section primarily presents the algorithmic components of DiagCoT. First, the overall three-stage algorithm of DiagCoT is illustrated in Algorithm~\ref{alg: Complete Algorithm}. Subsequently, the construction algorithm for the CoT dataset in the second stage of DiagCoT is described in Algorithm~\ref{alg: Stage2 Algorithm}.

\begin{algorithm}[ht]
\caption{Complete Algorithm}
\label{alg: Complete Algorithm}
\begin{algorithmic}[1]
  \State \textbf{Input:} Image captioning dataset $\mathcal{D}_\text{SFT} = \{(x_i,y_i)\}_{i=1}^K$, $\mathcal{D}_\text{RFT} = \{(x_i,y_i)\}_{i=1}^l$, Base VLM $\mathcal{M}_\text{base}$, expert VLM $\mathcal{M}_\text{expert}$;
  % search strategies $\mathcal{K}$, max search depth $N$, max search attempts $T$;

  \Statex \textbf{Stage 1: Alignment Stage}
  \State Initialize base VLM parameters $\theta$;
  \For{each $(x_i,y_i)\in \mathcal{D}_\text{SFT}$}
    \State $\mathcal{L}_\text{SFT}(\theta)\gets -\log \mathcal{M}_\text{base}^\theta(y_i\mid x_i)$;
    \State Update $\theta\gets\text{Optim}(\nabla_\theta \mathcal{L}_\text{SFT},\theta)$;
  \EndFor
  \State Obtain aligned model $\mathcal{M}_{\text{Stage1}}$;

  \Statex \textbf{Stage 2: CoT-Tuning Stage}
  \State Use Stage2 Algorithm Obtain $\mathcal{D}_\text{Final CoT}$;
  \For{each $(x,\tilde{e},\tilde{y})\in \mathcal{D}_\text{Final CoT}$}
    \State $\mathcal{L}_\text{SFT}(\theta)\gets -\log \mathcal{M}_\text{Stage1}^\theta(\tilde{e},\tilde{y} \mid x_i)$;
    \State Update $\theta\gets\text{Optim}(\nabla_\theta \mathcal{L}_\text{SFT},\theta)$;
  \EndFor
  \State Obtain aligned model $\mathcal{M}_{\text{Stage2}}$;
  \Statex \textbf{Stage 3: RFT-Tuning Stage}
  \State Use $\mathcal{J}_{\mathrm{GRPO}}(\theta)$ and $\mathcal{D}_\text{RFT}$ train $\mathcal{M}_{\text{Stage2}}$;
  \State Obtain $\mathcal{M}_{\text{Stage3}}$
\end{algorithmic}
\end{algorithm}

\begin{algorithm}[htbp!]
\caption{Stage2 Algorithm}
\label{alg: Stage2 Algorithm}%标签要紧跟caption之后
\begin{algorithmic}[1]

  \State \textbf{Input:} Image captioning dataset $\mathcal{D}_\text{SFT} = \{(x_i,y_i)\}_{i=1}^K$, Base VLM $\mathcal{M}_\text{base}$, expert VLM $\mathcal{M}_\text{expert}$, search strategies $\mathcal{K}$, max search depth $N$, max search attempts $T$;
  \Statex \textbf{Stage 2: CoT-Tuning Stage}
  \State \textbf{(1) Train Med-init $\mathcal{M}_{\mathrm{med init}}$}
  \For{each $(x_i,y_i)\in \mathcal{D}_\text{SFT}$}
    \State $\mathcal{L}_\text{SFT}(\theta)\gets -\log \mathcal{M}_{\text{med init}}^\theta(y_i\mid x_i)$;
    \State Update $\theta\gets\text{Optim}(\nabla_\theta \mathcal{L}_\text{SFT},\theta)$;
  \EndFor
  \State Obtain $\mathcal{M}_{\text{med init}}$

  \State \textbf{(2) CoT dataset collection}
  \State $\mathcal{D}_\text{random}\gets\text{RandomSelect}(\mathcal{D}_\text{SFT})$;
  \State $\mathcal{D}_\text{CoT}\gets \emptyset $;
  \For{each $(x,y)\in \mathcal{D}_{\text{random}}$}
    \For{$j\gets1$ \textbf{to} $T$}
      \State $(e_0,y_0) \gets \mathcal{M}_{\text{med init}}(X, Y)$;
      \For{$i\gets1$ \textbf{to} $N$}
        \State $c_i\sim\mathcal{C}$;
        \State $(e_i,y_i)\gets\mathcal{M}_{\text{med init}}^{c_i}\bigl(x,[e_0,y_0,\dots,e_{i-1},y_{i-1}]\bigr)$;
        \If{Verifier$(y_i,y^*)$}
          \State $\tilde{e}, \tilde{y} \gets \mathcal{M}_{\text{med init}}^{\mathrm{Reformat}}([e_0,y_0,\dots,e_i,y_i])$;
          % \State $\tilde{y} \gets \mathcal{M}_{\text{med init}}^{\mathrm{Response}}(\hat e)$;
          \State $\mathcal{D}_\text{CoT} \gets \{(x,\tilde{e},\tilde{y})\}$;
          \State \textbf{break};
        \EndIf
      \EndFor
      \If{Verifier$(y_i,y)$}
        \State \textbf{break};
      \EndIf
    \EndFor
  \EndFor
  \State Obtain $\mathcal{D}_\text{CoT}$;
  \State \textbf{(3) CoT Dataset Filtering}
  \State $\mathcal{D}_\text{Final CoT} \gets \emptyset $ 
  \For{each $(x,\tilde{e},\tilde{y})\in \mathcal{D}_{\text{CoT}}$}
    \State Flag $\gets \mathcal{M}_\text{expert}(x,\tilde{e},\tilde{y},y)$;
    \If{Flag == True}
        \State $\mathcal{D}_\text{Final CoT} \gets (x,\tilde{e},\tilde{y})$;
    \EndIf
  \EndFor
  \State Obtain $\mathcal{D}_\text{Final CoT}$;

\end{algorithmic}
\end{algorithm}

%\end{refsection}
\end{document}